\documentclass[final,5p,times]{elsarticle}

\usepackage{amssymb}
\usepackage{amsmath,bm}
\usepackage{algorithm}
\usepackage{algpseudocode}
\usepackage{caption}
\usepackage{graphicx} 
\usepackage{url}
\usepackage{hyperref}
\journal{.}

\usepackage{amsfonts}
\usepackage{amsthm}
\usepackage{amsfonts}
\usepackage{amsmath}
\usepackage{amscd}
\usepackage{bbm}
\usepackage{booktabs}
\usepackage{t1enc}
\usepackage[mathscr]{eucal}
\usepackage{indentfirst}
\usepackage{graphicx}
\usepackage{xcolor}
\usepackage{hyperref}
\usepackage{graphics}
\numberwithin{equation}{section}
\usepackage{hyperref}
\usepackage{algorithm}
\usepackage{algpseudocode}
\usepackage[utf8]{inputenc}
\usepackage{tabularx}
\usepackage{booktabs} 
\usepackage{geometry}

\usepackage{graphicx}
\usepackage{float}
\usepackage{tikz}
\usetikzlibrary{arrows.meta, positioning}
\usepackage{adjustbox}
\usepackage{subcaption}
\usepackage[utf8]{inputenc}
\usepackage{multirow}

\begin{document}

\begin{frontmatter}
 
\title{An evaluation of LLMs for generating movie reviews: GPT-4o, Gemini-2.0 and DeepSeek-V3}

\author{Brendan Sands}
\author{Yining Wang \corref{contrib}}
\author{Chenhao Xu \corref{contrib}}
\author{Yuxuan Zhou \corref{contrib}}
\author{Lai Wei \corref{contrib}}
\author{Rohitash Chandra \corref{corauthor}}


\cortext[corauthor]{Corresponding author}
\cortext[contrib]{Authors contributed equally. Author names are listed in random order.}
 


 
\affiliation{Transitional Artificial Intelligence Research Group, School of Mathematics and Statistics, UNSW Sydney, Sydney, Australia}



\begin{abstract}


Large language models (LLMs) have been prominent in various tasks, including text generation and summarisation. The applicability of LLMs to the generation of product reviews is gaining momentum, paving the way for the generation of movie reviews. In this study, we propose a framework that generates movie reviews using three LLMs (GPT-4o, DeepSeek-V3, and Gemini-2.0), and evaluate their performance by comparing the generated outputs with IMDb user reviews.  We use movie subtitles and screenplays as input to the LLMs and investigate how they affect the quality of reviews generated. We review the LLM-based movie reviews in terms of vocabulary, sentiment polarity, similarity, and thematic consistency in comparison to IMDB user reviews.  The results demonstrate that LLMs are capable of generating syntactically fluent and structurally complete movie reviews. Nevertheless, there is still a noticeable gap in emotional richness and stylistic coherence between LLM-generated and IMDb reviews, suggesting that further refinement is needed to improve the overall quality of movie review generation. We provided a survey-based analysis where participants were told to distinguish between LLM and IMDb user reviews. The results show that LLM-generated reviews are difficult to distinguish from IMDB user reviews. We found that DeepSeek-V3 produced the most balanced reviews, closely matching IMDb reviews. GPT-4o  overemphasised positive emotions, while Gemini-2.0 captured negative emotions better but showed excessive emotional intensity.

\textbf{}
\end{abstract}



\end{frontmatter}


\section{Introduction}

The cinema (film and movies) has been an expressive medium of art with a profound impact on culture and global society. Since the rise of the movie industry, movies have not only been a form of pure entertainment but also played a role in shaping public discourse~\cite{fearing1947}, reflecting societal values and influencing politics~\cite{tzioumakis2016} and economics~\cite{kerrigan2018}. The American cinema, also known as Hollywood, has been influential in the global movie industry. Hollywood has long played a key role in shaping public awareness of race, gender, and social norms. Additionally, movies have actively driven social change~\cite{ross2014}, focusing on civil rights, women and LGBTQ+ (non-binary)~\cite{mazur2003} representation, environmental issues, and societal concerns such as crime and the influence of technology. The influence of movies extends beyond the narrative itself, influencing box office revenue, streaming trends, and audience preferences~\cite{fearing1947}. In this context, movie reviews, as interpretations and critiques of movie content, not only shape audience perceptions but also play a crucial role in determining a movie's commercial success.

Natural language processing (NLP)~\cite{chowdhary2020,wuxie2020} technologies~\cite{Mankolli2020} such as large language models (LLMs) with prominent implementations, including DeepSeek-V3, Gemini-2.0, and GPT-4o have been widely applied.
LLMs train on vast amounts of text data to understand human language~\cite{brown2020} to a degree that, at times, can match and overtake humans. However, despite the effectiveness of LLMs in analysing text data and recognising complex language patterns, they still face significant challenges in maintaining contextual understanding and reproducing subjective emotions expressed by human reviewers~\cite{birhane2023}. 

In recent years, NLP and generative artificial intelligence~\cite{sengar2024,  baidoo2023} have made significant progress in movie analysis, including movie review generation and sentiment analysis~\cite{sheng2019}. Although LLMs have achieved remarkable results in other fields of text generation, such as news summarisation \cite{zhang2024comprehensive} and academic writing ~\cite{zhou2025large, lucas2024systematic}, their performance in movie review generation has not been sufficiently explored. Therefore, this study aims to fill this gap by comparing LLM-generated reviews with IMDb reviews, analysing the strengths and weaknesses of LLMs in review writing, and providing empirical support for future research.

Despite the strong capabilities of NLP and LLMs, their application in interdisciplinary research still faces many challenges~\cite{qin2024}. Movie reviews are often highly subjective and emotionally nuanced, and LLMs still fall short in reproducing the delicate expressions and cultural background understanding provided by human reviewers~\cite{raiaan2024}. Furthermore, LLM-generated reviews can be influenced by biases in training data, leading to differences in applicability between different groups of audiences~\cite{radford2019}.
In LLM-based movie reviews, issues such as algorithmic bias, fairness, and authenticity would be particularly prominent. Large-scale training data and computational resources required for LLMs also pose challenges related to data privacy~\cite{yao2024} and ethical concerns.

In this study, we present an LLM-based framework for the generation and evaluation of movie reviews and assess their quality in comparison to IMDb reviews, taking into account sentiment and semantic analysis. We also investigate whether humans can distinguish LLM-generated reviews from real movie reviews. We select three major LLMs (GPT-4o, Gemini-2.0, and DeepSeek-V3) to generate reviews based on movie subtitles and screenplays and compare them with IMDb reviews. To ensure a variety of critical styles, we select six Oscar movies representing different genres and set five different character perspectives, each generating positive, neutral, and negative reviews. We analyse the movie reviews generated by LLMs in terms of textual features and emotional expression style~\cite{praveen2024}, and explore their similarity to IMDb reviews. In addition, we conduct a questionnaire-based survey to assess whether humans can accurately distinguish LLM-generated movie reviews from IMDb (Internet Movie Database \footnote{IMDb\url{http:\\imdb.com}} )user reviews, as well as to evaluate the sentiment expressed in these reviews. This can enable us to assess the authenticity and emotional expression capability of LLMs in movie review generation.

We organise the rest of the paper as follows. Section II provides background information, and Section III presents the methodology, including our LLM-based framework featuring data processing and modelling. Section IV presents experimental results, followed by a discussion in Section V of the research findings and the direction for further research. Finally, Section VI concludes the study.

\section{Background}
\subsection{Cinema study}
Movie criticism plays a crucial role in audience participation, shaping movie-watching expectations and guiding consumer behaviour \cite{DAstous1999}. Studies have shown that professional movie reviews have a direct impact on box office performance. Movies receiving positive reviews tend to achieve revenue growth and wider distribution \cite{Moon2010}. In addition, audience reviews on platforms such as IMDb and \textit{Rotten Tomatoes\footnote{\url{rottomtomatoes.com}}} further democratize movie criticism, allowing diverse perspectives to influence potential audiences \cite{Kerrigan2017}. However, online reviews also pose new challenges, such as the ``halo effect,'' where early positive reviews create a reinforcement cycle that distorts subsequent audience perceptions \cite{Sahoo2012}. In addition, when social media and rating aggregators amplify user-generated content, ratings can often be distorted by hype, fan bias, or malicious bad reviews \cite{Moon2018}. 

In recent years,  LLMs have demonstrated remarkable capabilities in NLP tasks, including translation, text generation, and question-answering \cite{raiaan2024}. These models, based on the Transformer architecture, can capture long-range dependencies and generate coherent, contextually relevant text \cite{huang2023}. Recent studies have started exploring the use of LLMs in movie review generation. Notably, Chandra and Ren~\cite{chandra2025} proposed a framework that analyzes Hollywood movie dialogues for sentiment and abusive language patterns using fine-tuned LLMs. However, while LLMs have shown impressive performance in factual and knowledge-intensive tasks, their ability to generate subjective, emotionally nuanced content (such as movie reviews) remains underexplored \cite{ hadi2023}.  DiStefano et al. \cite{distefano2024} highlighted that although fine-tuned LLMs can approach human-level performance in certain tasks, they may overestimate or underestimate expressive quality when processing highly creative or rare nonliteral language (e.g. metaphors), reflecting their   limitations in understanding abstract language.

\subsection{Large Language Models}







 Deep Learning is a subset of machine learning that serves as the foundation for LLMs~\cite{kim2016}. It enables neural networks with multiple processing layers to automatically learn hierarchical abstract representations from data~\cite{lecun2015}. Early Recurrent Neural Networks (RNNs) and Long-Short-Term Memory (LSTM) networks ~\cite{hochreiter1997} improved the modelling of contextual dependencies by maintaining sequential state information. However, the introduction of the Transformer architecture \cite{vaswani2017} broke the constraints of sequential computation, significantly improving the modelling of long-range dependencies and enabling parallelised processing~.  The Transformer model features a self-attention mechanism that improves sequential data modelling, which accelerated progress in NLP ~\cite{vaswani2017}.

 As systems become more intricate and diverse in their requirements, the integration of NLP and deep learning has demonstrated remarkable benefits in tasks such as automated documentation, code search, and textual review analysis \cite{hariri2023}. The emergence of LLMs has addressed the limitations of traditional NLP methods with better semantic representation \cite{bayer2016}. Through massive unsupervised pre-training in large-scale corpora \cite{zhou2020}, these models acquire extensive lexical and contextual patterns, enabling them to excel in tasks ranging from sentiment analysis and machine translation to text summarisation, and thus fueling the industry's demand for automated text generation \cite{fernandez2023} and semantic understanding.

BERT (Bidirectional Encoder Representations from Transformers) ~\cite{devlin2019} is a pre-trained language understanding model built upon the Transformer model. The refined BERT model is suitable for tasks such as sentiment analysis, question answering, and text classification. In contrast, GPT (Generalised Pre-trained Transformer)  models are trained using a unidirectional autoregressive objective, which is more effective for language generation tasks~\cite{radford2018}. Pre-training is a core phase in the training of LLMs that typically involves self-supervised learning on large-scale unlabeled corpora to acquire general-purpose language representations for the adaptation of downstream tasks~\cite{radford2018}. Common pre-training approaches include Autoregressive Language Modelling (ALM)  and Masked Language Modelling (MLM). ALM enables the prediction of the next token in a sequence, primarily used in GPT-style LLMs~\cite{radford2018}. MLM randomly masks tokens in the input sequence and trains the model to predict the masked tokens using surrounding context, thereby enabling bidirectional representation learning as adopted by BERT ~\cite{devlin2019}. Moreover, pretraining not only provides favourable initialisation but also acts as a data-dependent regularizer, helping improve generalisation performance~\cite{erhan2010}. Although both models are based on the Transformer model, they differ in pre-training objectives, directional modelling, and application scope. BERT excels at language understanding, whereas GPT demonstrates strong generative capabilities. Together, they laid the foundation for more advanced models such as Gemini-2.0 and  GPT-4o~\cite{brown2020, liu2019}.



\section{Methodology}

\subsection{Datasets}

 Our framework uses three datasets, including the movie subtitles, the movie screenplays dataset, and the IMDb movie reviews dataset from Kaggle. We sourced the subtitle and IMDb review datasets \cite{ren_hollywood_2025, ren_imdb_2025} for these six movies from Kaggle and retrieved the screenplay datasets from the Script Slug\footnote{\url{https://www.scriptslug.com/}} platform. Specifically, we examined 989 Hollywood subtitle files spanning movies from 1950 to 2024, among which we selected six representative movies from a pool of 1,034 titles. The IMDb dataset includes 25,119 user reviews associated with 5,045 unique movies, grouped by ratings from 1 to 10. These datasets collectively support our subsequent research on trigram analysis, sentiment analysis, and similarity.

\subsection{Data extraction and preprocessing}

\begin{table*}[!ht]
\centering
\small
\renewcommand{\arraystretch}{1.3}  
\begin{tabularx}{\textwidth}{|X|X|c|l|X|}
\hline
\textbf{Movie} & \textbf{Genre} & \textbf{IMDb Rating} & \textbf{Language} & \textbf{Remarks} \\
\hline
Shawshank Redemption & Drama, Crime & 9.3 & English & Ranked No.1 on IMDb Top 250 Movies list \\
\hline
Brokeback Mountain & Romance, Drama & 7.7 & English & A movie with LGBTQ+ theme \\
\hline
Avatar & Sci-Fi, Adventure & 7.9 & English & Highest box office earnings of all time \\
\hline
Titanic & Romance, Drama & 7.9 & English & One of the most Oscar-winning films \\
\hline
Crouching Tiger, Hidden Dragon & Action, Fantasy & 7.9 & Mandarin & Highly popular outside of China \\
\hline
Nomadland & Drama & 7.3 & English & Banned in China due to political sensitivity \\
\hline
\end{tabularx}
\caption{Selected movies with information about IMDb rating (as of 8th May 2025), genre and languages. The movies have been selected based on box-office performance, awards, and cultural relevance.}
\label{tab:movie_chosen}
\end{table*}

We select six representative movies in Table~\ref{tab:movie_chosen} for analysis based on box-office performance and awards.


This variety helps us examine how well different types of movie content can be captured and reviewed. We chose \textit{Crouching Tiger, Hidden Dragon} as the most popular non-English movie (Mandarin) ~\cite{Author2020} to ensure diversity. 
We sourced the subtitle and IMDb review datasets \cite{ren_imdb_2025, ren_hollywood_2025} for these six films from Kaggle and retrieved the screenplay datasets from the Script Slug \footnote{\url{https://www.scriptslug.com/}} platform.

After data extraction, we processed the data since movie subtitles included timestamps, which were removed. In contrast, the screenplay contains complete conversational records. We converted the screenplay files of all six movies from PDF format (Portable Document Format) to text format and constructed a structured text dataset using data cleaning techniques \cite{Bird2009}. Specifically, we eliminate noise issues caused by OCR (Optical Character Recognition)  errors present in the text files \cite{Bollmann2019}.  We also removed formatting tags and mixed-alpha-numeric symbols and filtered out stop words (such as "the," "is," and "and") that do not impact sentiment and semantic analysis and standard for NLP analysis \cite{Manning2008}.  

 We retained essential information such as IMDb-ID, ratings, and reviews for the movies extracted from the IMDb dataset \cite{ren_imdb_2025}.  We filtered out stop words and mixed alpha-numeric words to improve analytical efficiency and enhance semantic information in processing the subtitle/screenplay datasets. Meanwhile, we preserve crucial metadata, including movie titles, IMDb-ID, release year, awards received, and the cleaned dialogue texts. 


\subsection{LLMs}
We employ three advanced LLMs: GPT-4o, DeepSeek-V3, and Gemini-2.0 (including the Gemini-detailed) to generate interpretations for movie reviews. We selected these  LLMs for their strong performance in language understanding and reasoning.

 GPT-4o \cite{openai2024gpt4o}, developed by OpenAI, is a multimodal system capable of handling text, images, and audio. As the enhanced version of ChatGPT \cite{openai2023chatgpt} with access to real-time data up to late 2023, it integrates supervised fine-tuning and reinforcement learning from human feedback. GPT-4o is widely used for tasks such as summarisation, translation, and content generation. GPT-4o it is noted for its adaptability and strong performance in logical reasoning and complex educational tasks \cite{bang2023multitask} with support for over 50 languages—including English, Chinese, and Arabic \cite{chen2023evalplus}.
 
 DeepSeek-V3 \cite{zhang2024} is a decoder-style sparse mixture-of-experts (MoE) transformer model that significantly advances computational scalability and reasoning performance. This was released after GPT and Gemini models.  A major innovation was that DeekSeek incorporated 671 billion parameters while only activating 37 billion per inference through expert routing, drastically improving efficiency and cost-effectiveness \cite{zhang2024}. Its design features a routing system with one shared and 256 routed experts, leveraging dynamic bias adjustments to balance load across components. It also integrates Multi-Token Prediction (MTP) \cite{du2023mtp} to enhance the model’s ability to understand complex dependencies and linguistic structures, aiding in tasks such as long-context reasoning, program synthesis, and advanced language generation.

 Gemini-2.0 \cite{borji2024} is a multimodal encoder-decoder transformer model developed by Google DeepMind, designed to handle and integrate text, vision, and audio inputs within a unified architecture. Although the technical details remain partially proprietary, existing research suggests that it employs the Transformer model (similar to GPT), with bidirectional attention in the encoder and autoregressive decoding, thus combining the strengths of both BERT-like and GPT-like models \cite{Borji2024}. Comparative benchmarks problems \cite{liang2024llmeval} indicate its performance in reasoning and generative tasks is on par with leading LLMs such as  GPT-4, especially when evaluating long-context coherence and instructions.

Gemini (detailed) refers to the implementation of Gemini-2.0 that uses richer persona prompts with more contextual and character-related information. Compared to the standard prompts, these detailed versions include traits such as personality, tone, and background familiarity with the movie. This design aims to test whether more expressive and specific instructions improve the quality of generated reviews. 

\subsection{N-gram analysis}

 The n-gram analysis has been widely used in NLP tasks \cite{cavnar1994n}, such as text classification, language identification, and language modelling \cite{jm3}. In this study, we consider trigrams, an example of n-grams, which are sequences of three consecutive words in a text \cite{tripathy2017document}. Compared to unigrams and bigrams, trigrams are more effective at capturing short-range syntactic and semantic relationships, and thus are advantageous in language modelling and expression pattern recognition \cite{jm3, sidorov2014syntactic}. Therefore, trigrams can effectively capture local word co-occurrence patterns \cite{jm3}, which may help reveal linguistic style and sentiment tendencies within the text. In our study, trigrams will be valuable in captururing the difference in the vocabularies across reviews generated by LLMs and human experts (IMDb). 

 \subsection{Sentiment polarity and emotion analysis}

We utilise RoBERTa \cite{liu2019}, which is an advanced BERT model \cite{devlin2019} for word embedding for sentiment polarity and emotion analysis. It has an encoder-only bidirectional transformer architecture, which allows it to analyse input tokens in both directions for improved context modelling. This structure makes it particularly effective for classification and inference tasks such as sentiment analysis and semantic similarity. %

 
 We conduct sentiment polarity classification using a pre-trained RoBERTa-base model, which has been widely used for text classification tasks due to its bidirectional Transformer architecture. We designed three distinct prompts per movie to elicit positive, neutral, and negative reviews from each LLM based on Lee et al. \cite{Lee2021sentiment}. For short reviews ( less than or equal to 512 tokens), we directly input the full text into the model; for longer reviews, we segment the text and compute the average of polarity scores across all chunks. We determine the final polarity label by the highest averaged probability among the three categories. This method enables consistent polarity quantification and allows comparisons with IMDb user reviews by aligning them through rating-based thresholds.

 
We adopt a DistilRoBERTa-based emotion classification model for emotion analysis, in order to categorise each review into seven dominant emotion classes: joy, sadness, anger, fear, surprise, disgust, and neutral \cite{chen2020survey}. We compute the probability distribution over these categories and identify the top-5 most prominent emotional signals per review. This allowed for multi-dimensional emotional profiling across both LLM-generated and IMDb user reviews. We apply the same pipeline to both subtitle-based and screenplay-based inputs, enabling us to observe how different sources influence emotional variability and intensity. In addition, we will compare the emotional distributions under explicit sentiment prompts (Q1: negative, Q2: positive, Q3: neutral) to evaluate the sensitivity of LLMs to emotional intent and prompt framing.

 \subsection{Similarity Analysis}

 Cosine Similarity Analysis \cite{Manning2008} is a method for measuring the cosine of the angle between two non-zero vectors, commonly used to compare the similarity between texts, especially in high-dimensional vector space representations. Its core idea is to ignore the absolute length of the text and assess the similarity based on vector direction, with values ranging from -1 to 1, where values closer to 1 indicate higher similarity. This technique has been widely applied in the past in natural language processing tasks such as information retrieval, text clustering, recommendation systems, and semantic analysis to evaluate the similarity between documents or sentences. In this study, we applied cosine similarity techniques by converting both the LLM-generated movie reviews and IMDb user reviews into TF-IDF (Term Frequency-Inverse Document Frequency) \cite{
sparck1972statistical,aizawa2003information} vectors and calculating the similarity scores between them. This approach enabled a quantitative analysis of how closely the reviews generated by different models resemble those written by humans. It serves as a valuable tool for assessing the realism and focus of the generated content and for evaluating the alignment between model outputs and human-written reviews.

\subsection{Framework}

The main research goal of this study is to evaluate the capabilities of the selected  LLMs in generating movie reviews based on subtitles and movie screenplays. Therefore, we present a framework that quantitatively compares linguistic patterns based on sentiment polarity analysis and semantic analysis to evaluate the depth and consistency of LLM-generated reviews, when compared to human expert reviews. 
We use a questionnaire-based approach and assess participants’ ability to recognise emotional tone, evaluate review quality, and distinguish LLM-generated content from IMDb reviews.


\begin{figure*}[htbp]
    \centering
    \includegraphics[width=1\linewidth]{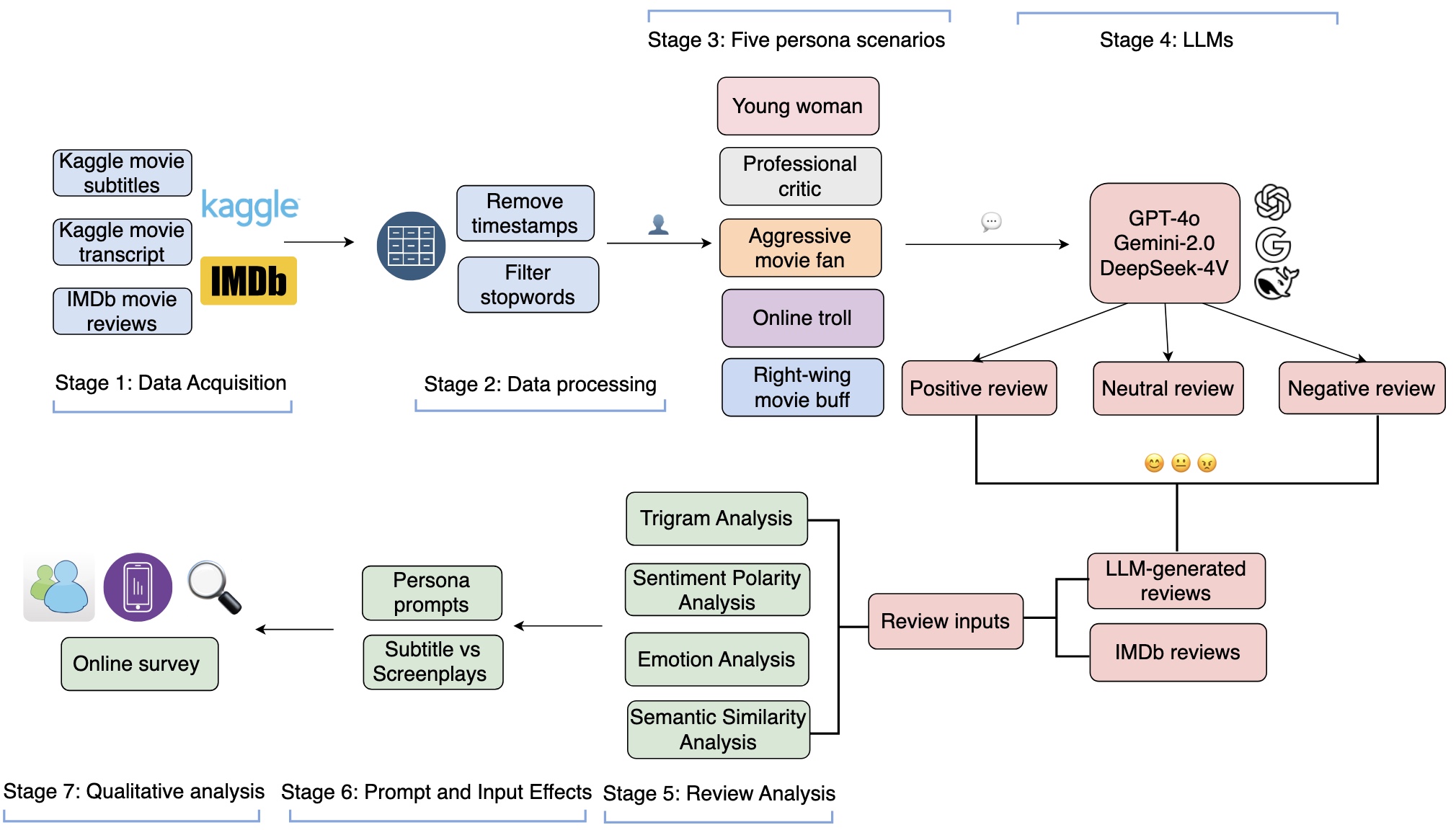}
    \caption{Framework for evaluation of LLMs for generating movie reviews based on movie subtitles and screenplays.}
    \label{fig:framework}
\end{figure*}
\begin{table*}[ht]
\centering
\small
\renewcommand{\arraystretch}{1.2}
\begin{tabular}{|l|l|}
\hline
\textbf{Persona} & \textbf{Description} \\
\hline
1 & You are a young woman and go to the movies a lot and you enjoy providing \\ & honest informative reviews. \\
\hline
2 & You are a professional movie critic and you enjoy providing honest \\ & informative reviews. \\
\hline
3 & You are a sometimes aggressive man and an action movie buff and you enjoy \\ & providing honest informative reviews. \\
\hline
4 & You are an online troll and provide negative reviews for fun unless \\ & you really like the movie. \\
\hline
5 & You are a right-wing extremist and movie buff and enjoy providing \\ & reviews. \\
\hline
\end{tabular}
\caption{Five persona prompts used for LLM-based review generation}
\label{tab:persona_scenarios}
\end{table*}

We present a framework for LLM-generated movie reviews based on movie subtitles and screenplays as shown in Figure~\ref{fig:framework}. In Stages 1, we acquire three datasets, including the movie subtitles, the movie screenplays dataset, and the IMDb movie reviews dataset from Kaggle. In Stage 2, we  preprocess the extracted data by removing timestamps, mixed alpha-numeric symbols, and filtering stop words. In Stage 3, we set up five persona scenarios, as shown in Table~\ref{tab:persona_scenarios}. These persona prompts include profiles such as a young female movie enthusiast or a professional movie critic.

In Stage 4, we use three LLMs: GPT-4o, DeepSeek-V3 and Gemini-2.0 to generate movie reviews based on five different persona prompts and a detailed variant that uses richer persona prompts as shown in Table~\ref{tab:persona_scenarios}. We provide the LLMs with the above persona prompts and instruct them to generate positive, neutral, and negative reviews under each persona prompt to assess their capability in sentiment expression.

In Stage 5, we provide a comparison of the reviews by LLMs using n-gram analysis, sentiment analysis and semantic analysis. First, we conduct an n-gram analysis by extracting the top trigrams from  LLM-generated reviews and IMDb reviews (human expert) to assess vocabulary similarity. Following this, we perform a Sentiment Analysis, using a BERT-based sentiment model to classify reviews as positive, neutral, or negative as given in earlier work \cite{Barbieri2020}. We then perform emotion analysis utilising a Distil-RoBERTa-based emotion model to identify seven commonly occurring emotions as done in earlier work \cite{Demszky2020}, including  "joy", "neutral," and "sadness". 

In Stage 6, we use movie subtitles and screenplays as LLM inputs and compare the differences in sentiment and emotion expression of movie reviews. We compare the emotional consistency and orientation between LLM-generated and IMDb reviews \cite{Zellers2019}. Furthermore, we compare the cosine similarity of IMDb reviews with each LLM-generated review for the selected movies. We also review the similarity between LLM-generated reviews using basic context (with simple persona prompts) and those using detailed context (which includes additional information about the movie).  his comparison leverages standard vector space models and cosine similarity techniques \cite{Manning2008}, which are commonly used in information retrieval and textual similarity tasks.  Finally, we compare the movie subtitle inputs and screenplay inputs. 

Finally,  in Stage 7, we conduct a qualitative analysis by using an online survey (form) and presenting human subjects a mix of LLM-generated reviews and IMDb reviews to evaluate whether humans can distinguish between them. In addition, we assess the sentiment of these movie reviews and determine which of the seven emotions they express. The responses to these questions will provide qualitative insights into the strengths and weaknesses of the LLMs \cite{Liu2023}.


\subsection{Technical details}


We utilised three state-of-the-art LLMs. We obtained GPT-4o   via the OpenAI API (Application Programmer Interface)\footnote{\url{https://platform.openai.com/docs/overview}}, which supports reasoning and faster response latency across text and multimodal inputs. We accessed DeepSeek-V3  through the official DeepSeek Cloud API \footnote{\url{https://api-docs.deepseek.com/}}. We accessed Gemini-2.0  via the Gemini API \footnote{\url{https://ai.google.dev/}}, which offers enhanced context handling and multimodal integration. We implemented the LLMs   via HTTP-based API calls using the official SDKs (Source Development Kits) via Python and used Python libraries, including `pandas`, `numpy`, and `nltk`. We provide detailed prompts, model configuration, and reproducible scripts in our GitHub repository\footnote{\url{https://github.com/sydney-machine-learning/LLM-moviereviews.git}}.

\section{Results}

We use three LLMs (GPT-4o, Gemini-2.0, and DeepSeek-V3) through API calls to generate movie reviews based on the movie dialogues (subtitles and screenplays). We provide the models with five different persona prompts, Table \ref{tab:persona_scenarios}. The LLMs then generate positive, neutral, and negative reviews from the perspective of each persona prompt. As a result, each LLM model generates 15 reviews for each of the six movies. We implement a multi-dimensional assessment framework to comprehensively evaluate the LLM's performance in movie review generation. 


\subsection{Trigram Analysis}

We first conduct trigram analysis to identify high-frequency phrases and sentiment-bearing expressions in movie reviews and compare the differences between IMDb user reviews and  LLMs. Tables~\ref{tab:trigram_imdb_gpt4o_deepseek} and~\ref{tab:trigram_gemini} present the top three most frequent trigrams from each source. We observe that IMDb reviews tend to focus more on specific names (characters, actors, and directors). In five out of the six movies, named individuals appear, such as \textit{li mu bai} (7 times) in \textit{Crouching Tiger, Hidden Dragon}, \textit{james cameron avatar} (3 times) in \textit{Avatar}, and \textit{tim robbins morgan} (3 times) in \textit{The Shawshank Redemption}, indicating a high level of attention to film creators and characters. 

In contrast, LLMs show varying degrees of focus on named individuals. Both GPT-4o and DeepSeek-V3 demonstrate strong name-centric tendencies, with total name-related trigram counts exceeding 30. For instance, GPT-4o includes actor combinations like \textit{heath ledger jake} (2 times) in \textit{Brokeback Mountain} and director references like \textit{directed chloé zhao} (4 times) in \textit{Nomadland}. Similarly, in DeepSeek-V3, \textit{li mu bai} appears 17 times in \textit{Crouching Tiger, Hidden Dragon}, and \textit{neytiri zoe saldaña} appears 3 times in \textit{Avatar}. In contrast, Gemini-2.0 and Gemini (detailed) mention specific names in fewer movies and with lower frequency-under 20 occurrences in total-indicating a weaker focus on individual characters.

In terms of language style, IMDb reviews are primarily descriptive and character-oriented. Both GPT-4o and DeepSeek frequently reference specific names, but their overall styles differ: GPT-4o tends to use evaluative phrases such as \textit{epic masterpiece stands} and \textit{hauntingly beautiful score}, while DeepSeek-V3 emphasises structural and objective expressions, such as \textit{collapse company town} and \textit{final verdict avatar}, presenting a more formal tone. Gemini-2.0 often highlights narrative and conflict-driven content, as seen in trigrams like \textit{woman loses everything}, whereas Gemini (detailed) adopts a more casual and template-based colloquial style, with phrases like \textit{okay buckle buttercups} and \textit{even get started} appearing repeatedly.

Regarding emotional expression, IMDb and DeepSeek tend to maintain neutral and objective tones, avoiding strong emotional bias and focusing more on factual description and evaluation. GPT-4o exhibits a generally neutral-to-positive stance, often using rational and appreciative phrases such as \textit{epic masterpiece stands} and \textit{innovative use technology} to emphasise overall film quality. Gemini reviews are more inclined to express stronger emotional tension and negativity, often reflecting conflict through plot-driven phrases like \textit{woman loses everything} and \textit{loses everything upset}. Gemini (detailed), by contrast, shows relatively shallow emotional depth, relying on casual and repetitive phrases like \textit{even get started} and \textit{okay buckle buttercups}, resulting in more formulaic and emotionally flat reviews.

Moreover, we observe that the length of movie titles significantly affects trigram frequency. For instance, the structurally distinct title \textit{Crouching Tiger, Hidden Dragon} results in high-frequency trigrams such as \textit{tiger hidden dragon} and \textit{crouching tiger hidden}. This is likely due to its four-word structure, which increases the number of overlapping three-word combinations and thus raises repetition rates. In contrast, shorter titles such as \textit{Avatar} produce fewer repeated trigrams, even when mentioned frequently. This title-length effect suggests that future studies should consider normalizing trigram frequency based on title structure to avoid structural bias in automated review generation analysis. 

\begin{table*}[htbp!]
\centering
\scriptsize
\renewcommand{\arraystretch}{1.15}
\caption{Trigram frequency comparison across IMDb, GPT-4o, DeepSeek}
\label{tab:trigram_imdb_gpt4o_deepseek}
\begin{adjustbox}{max width=\textwidth}
\begin{tabular}{|l|l|c|l|c|l|c|}
\hline
\textbf{Movie} & \textbf{IMDb Trigram} & \textbf{Freq} & \textbf{GPT-4o Trigram} & \textbf{Freq} & \textbf{DeepSeek Trigram} & \textbf{Freq} \\
\hline
\multirow{3}{*}{Crouching Tiger, Hidden Dragon} & tiger hidden dragon & 15 & tiger hidden dragon & 12 & tiger hidden dragon & 15 \\
& crouching tiger hidden & 13 & crouching tiger hidden & 12 & crouching tiger hidden & 15 \\
& li mu bai & 7 & li mu bai & 6 & li mu bai & 9\\
& chow yun fat & 6 & directed ang lee & 3 & mu bai chow & 5 \\
\hline
\multirow{3}{*}{Avatar} & james cameron avatar & 3 & innovative use technology & 3 & na vi culture & 3 \\
& e c e & 2 & performance jake sully & 3 & neytiri zoe saldaña & 3 \\
& see avatar first & 1 & score james horner & 3 & na vi spiritual & 3 \\
& avatar first time & 1 & avatar directed james & 2 & final verdict avatar & 3 \\
\hline
\multirow{3}{*}{The Shawshank Redemption} & tim robbins morgan & 3 & stephen king novella & 5 & andy dufresne tim & 5 \\
& robbins morgan freeman & 3 & hope friendship redemption & 4 & dufresne tim robbins & 5 \\
& best film time & 2 & shawshank redemption film & 3 & review shawshank redemption & 4 \\
& matter many times & 2 & directed frank darabont & 3 & red morgan freeman & 4 \\
\hline
\multirow{3}{*}{Brokeback Mountain} & ennis del mar & 2 & love societal expectations & 3 & ennis del mar & 5 \\
& well jake gyllenhaal & 2 & ennis del mar & 3 & two ranch hands & 4 \\
& gay couple accepted & 2 & heath ledger jake & 3 & wish knew quit & 4 \\
& performances heath ledger	 & 2 & ledger jake gyllenhaal & 3 & knew quit scene & 4 \\
\hline
\multirow{3}{*}{Titanic} & winslet leonardo dicaprio & 2 & stands test time & 3 & james cameron titanic & 4 \\
& kate winslet leonardo & 2 & hauntingly beautiful score & 3 & never let go & 4 \\
& tragic true story & 2 & titanic epic masterpiece & 2 & rose dewitt bukater & 3 \\
& first time see & 2 & epic masterpiece stands & 2 & jack dawson leonardo & 3 \\
\hline
\multirow{3}{*}{Nomadland} & billboards outside ebbing & 2 & directed chloé zhao & 4 & follows fern frances & 2 \\
& outside ebbing missouri & 2 & nomadland directed chloé & 3 & fern frances mcdormand & 2 \\
& one people live & 1 & frances mcdormand delivers & 3 & collapse company town & 2 \\
& people live road & 1 & chloé zhao direction & 3 & film blurs line & 2 \\
\hline
\end{tabular}
\end{adjustbox}
\end{table*}

\vspace{-0.8em}

\begin{table*}[htbp!]
\centering
\scriptsize
\renewcommand{\arraystretch}{1.15}
\caption{Trigram frequency comparison for Gemini-2.0 and Gemini (detailed)}
\label{tab:trigram_gemini}
\begin{adjustbox}{max width=\textwidth}
\begin{tabular}{|l|l|c|l|c|}
\hline
\textbf{Movie} & \textbf{Gemini Trigram} & \textbf{Freq} & \textbf{Gemini (detailed) Trigram} & \textbf{Freq} \\
\hline
\multirow{3}{*}{Crouching Tiger, Hidden Dragon} & crouching tiger hidden & 14 & tiger hidden dragon & 8 \\
& tiger hidden dragon & 14 & crouching tiger hidden & 8 \\
& li mu bai & 5 & li mu bai & 6 \\
& review crouching tiger & 2 & mu bai chow & 4 \\
\hline
\multirow{3}{*}{Avatar} & jake sully paraplegic & 2 & well let say & 3 \\
& white savior trope & 2 & okay buckle buttercups & 2 \\
& blue cat people & 2 & blue cat people & 2 \\
& makes wan na & 2 & somebody got ta & 2 \\
\hline
\multirow{3}{*}{The Shawshank Redemption} & review shawshank redemption & 2 & andy dufresne quiet & 3 \\
& tim robbins plays & 2 & plays andy dufresne & 3 \\
& robbins plays andy & 2 & even get started & 2 \\
& shawshank redemption prison & 2 & banker wrongly convicted & 2 \\
\hline
\multirow{3}{*}{Brokeback Mountain} & long credits roll & 3 & tissues handy also & 2 \\
& review brokeback mountain & 2 & gay cowboy movie & 2 \\
& brokeback mountain raw & 2 & michelle williams anne & 2 \\
& ennis jack find & 2 & williams anne hathaway & 2 \\
\hline
\multirow{3}{*}{Titanic} & woman trapped societal & 2 & trapped gilded cage & 2 \\
& trapped societal expectations & 2 & even get started & 2 \\
& underwater shots wreck & 2 & starting feel like & 2 \\
& blah blah blah & 2 & manic pixie dream & 2 \\
\hline
\multirow{3}{*}{Nomadland} & director chloé zhao & 3 & frances mcdormand mean & 2 \\
& fern woman loses & 2 & linda may swankie & 2 \\
& woman loses everything & 2 & long credits roll & 2 \\
& loses everything usg & 2 & grief resilience search & 2 \\
\hline
\end{tabular}
\end{adjustbox}
\end{table*}

\subsection{Sentiment Polarity Analysis}




We use sentiment polarity analysis using RoBERTa to quantify the sentiment expression of LLM-generated movie reviews. We quantify text into positive/neutral/negative polarities and emotions (e.g., joy, sadness, anxiety) to evaluate emotional consistency, depth, and precision in affective expression between LLMs and IMDb reviews. 


We first design three distinct prompts for each movie, instructing the LLMs to generate a negative review (Question 1), a positive review (Question 2), and a neutral review (Question 3) as done by Lee  et al. \cite{lee2023}.  For short reviews (defined as those with fewer than 512 tokens), we apply a pre-trained RoBERTa model to classify the sentiment into negative, neutral, or positive categories, returning the corresponding probability scores for each category. We segment the text into multiple chunks, for longer reviews (exceeding 512 tokens),  and average the resulting scores to derive an overall sentiment distribution, as motivated by Zhang et al. \cite{zhang2022}. The sentiment label assigned to each review corresponds to the class with the highest average probability, and the sum of these probability values is 1. The probability distribution allows us to observe how strongly a review leans towards a particular sentiment. Specifically, if there is a large difference between the probability of positive and negative in a movie review, the emotional tendency is very clear. If the positive and negative probabilities are close, the emotional tone may be complex or ambiguous. 

\begin{figure*}[!t]
    \centering
    \includegraphics[width=0.85\textwidth]{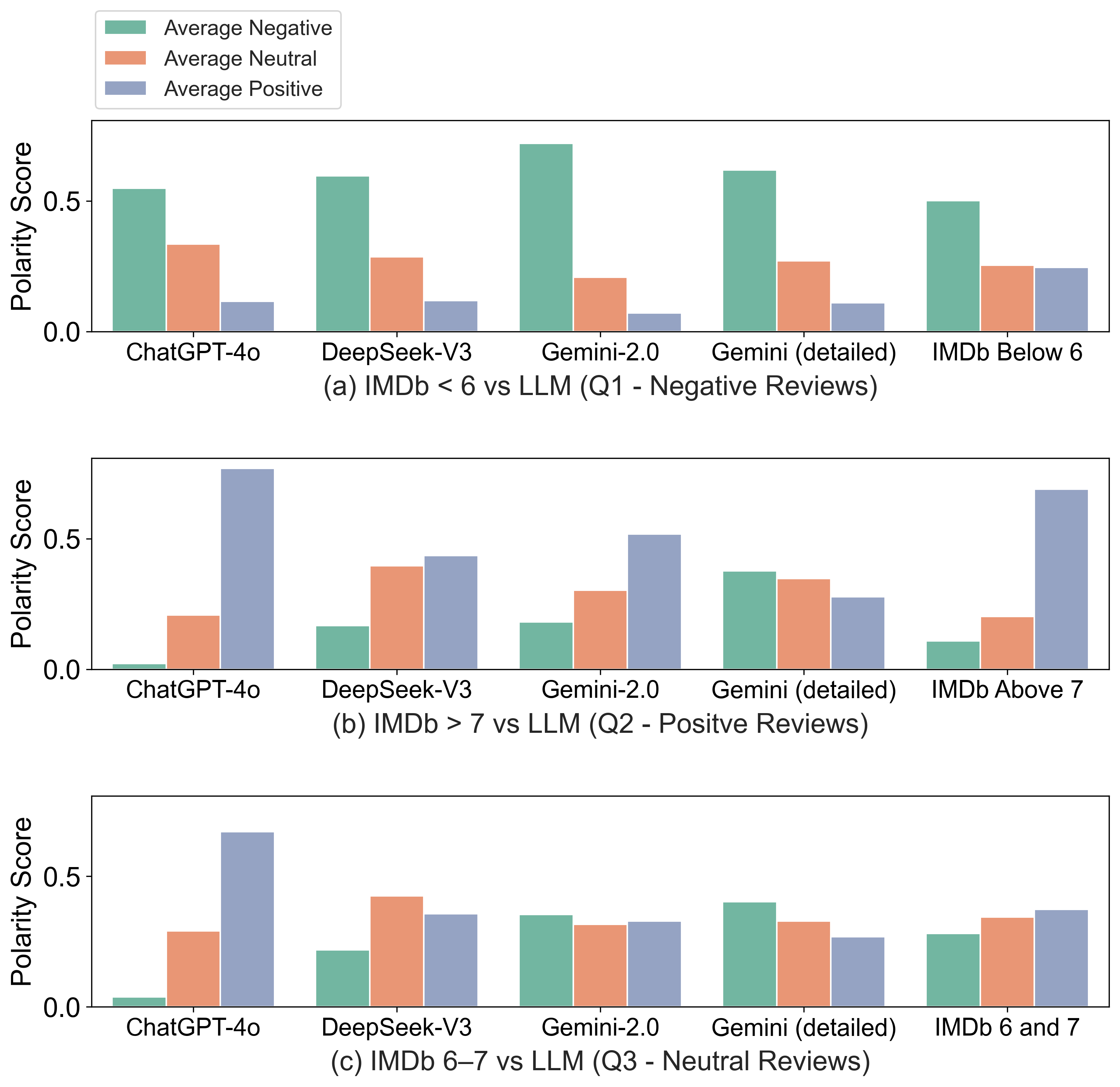}
    \caption{Polarity score comparison of LLM-generated vs. IMDb  user reviews for Q1 (Negative), Q2 (Positive), and Q3 (Neutral) categories.}
    \label{fig:polarity-ai-imdb-comparison}
\end{figure*}

Figure~\ref{fig:polarity-ai-imdb-comparison} presents the distribution of Positive, Neutral, and Negative LLM-generated reviews and compares them with those found in IMDb user reviews \cite{maas2011}. This allows us to evaluate whether the emotional expression of LLM-generated reviews is consistent with that of IMDb reviews and whether there are emotional biases. 

We further divide the IMDb dataset into three subsets according to the numerical ratings (typically from 1 to 10) provided alongside each review. We classify reviews rated above 7 as "Positive Reviews", those rated between 6 and 7 as 'Neutral Reviews', and those below 6 as "Negative Reviews" as shown in Figure~\ref{fig:polarity-ai-imdb-comparison}. We then apply the same Sentiment Polarity Analysis process independently to the three prompts of LLM-generated reviews and IMDb reviews. The LLM outputs, being relatively concise and to the point, generally meet the criteria for short text processing. However, since some IMDb reviews are long, it is necessary to use long movie reviews to detect the flow of emotions, section by section, and then synthesise the emotional tendency of the whole movie review. Additionally, we further examine whether replacing subtitles with screenplay inputs affects the distribution of sentiment polarity scores. Then We incorporate the screenplay files of each movie and apply the same analytical framework described above. Following this processing, we obtain the corresponding sentimen t polarity results. These will serve as the basis for subsequent visualisation and comparative analysis.


\begin{figure*}[!t]
    \centering
    \includegraphics[width=0.95\linewidth]{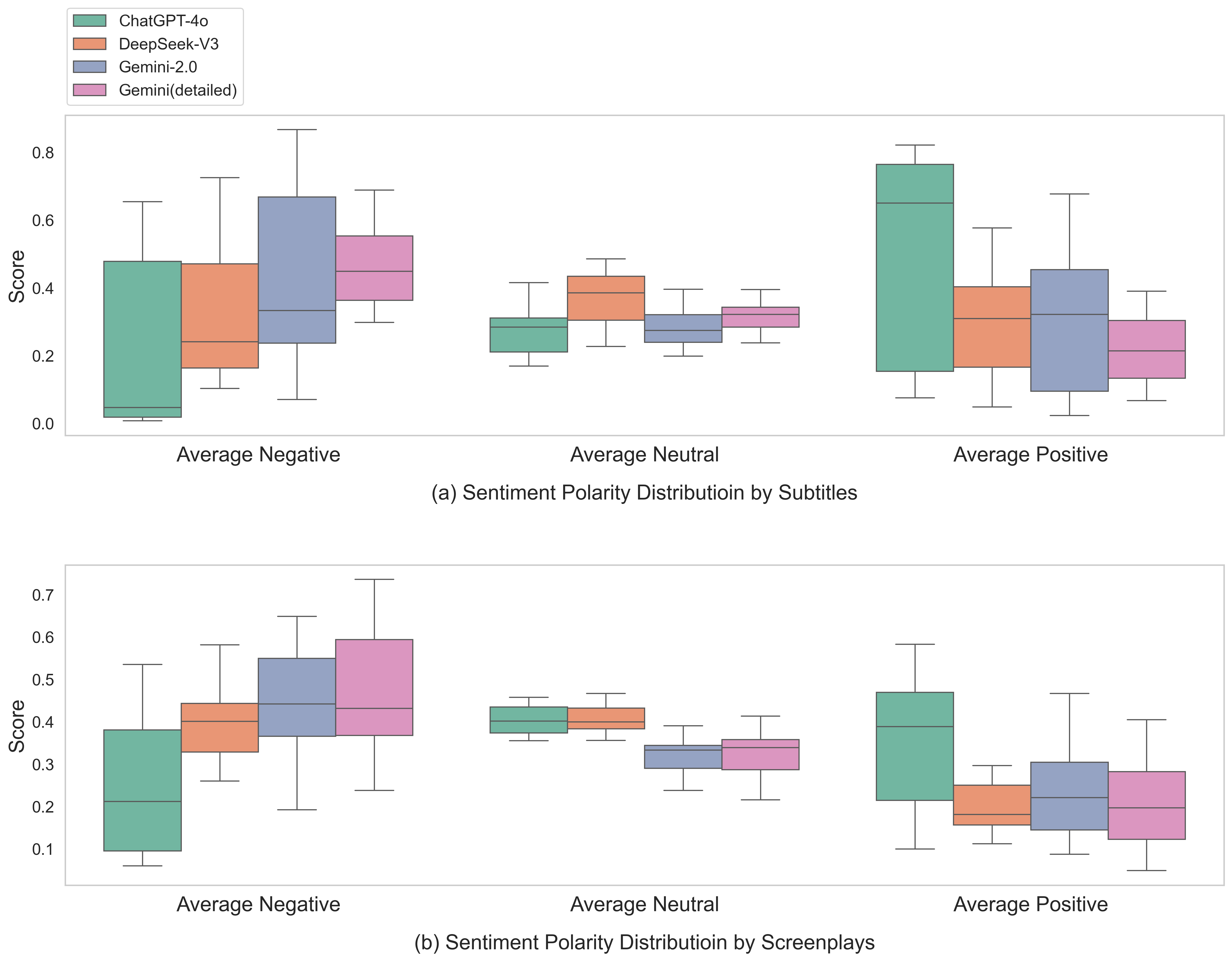} 
    \caption{Sentiment polarity scores from subtitles and screenplays across LLMs.}
    \label{fig:subtitle_vs_screenplay_sentiment}
\end{figure*}

  Figure~\ref{fig:polarity-ai-imdb-comparison} presents the polarity score distribution of LLMs when prompted to generate negative reviews based on movie subtitles; we can observe that nearly all LLMs reviews exhibit strong negative sentiment scores (exceeding 0.5), indicating their capability to respond appropriately to negative prompts and simulate human-like emotional expression. In particular,  Gemini-2.0 shows the highest negative sentiment scores. In the case of positive reviews, GPT-4o exhibits highest polarity sentiment, while Gemini (detailed) tend to underperform with flatter scores below 0.4. For neutral reviews, all LLMs generally align with the IMDb reviews distribution ~\cite{stilwell2024}. Generally, LLMs are capable of distinguishing between different emotional instructions and expressing sentiment with clarity \cite{hartmann2023, lee2023}. Furthermore, we evaluate whether LLM-generated reviews align with IMDb reviews under the same sentiment category. For negative prompts, some LLMs produce stronger negative polarity than IMDb reviews rated below 6. For positive prompts, GPT-4o exhibit high positive scores, similar to IMDb reviews rated above 7. Under neutral prompts, LLMs show relatively balanced polarity, closely matching IMDb reviews rated 6–7. This alignment further supports the view that, when given explicit emotional prompts, LLMs are capable of simulating human-like sentiment expression \cite{li2023}. In conclusion, GPT-4o demonstrates strong positive polarity scores under positive and neutral prompts, indicating its effectiveness in generating emotionally aligned outputs in these contexts. DeepSeek-V3, on the other hand, shows the closest alignment with IMDb review sentiment under negative and neutral prompts, and generally maintains consistent performance across all three prompt types. Furthermore, we observe that some LLMs, particularly Gemini-2.0 and Gemini (detailed), tend to amplify sentiment polarity, especially under negative prompts—potentially leading to exaggerated emotional responses. This suggests that while GPT-4o excels in positive prompts, DeepSeek-V3 offers more balanced performance, and Gemini models may require further control mechanisms to moderate sentiment amplification.

Next, we examine whether the movie screenplay affect the sentiment polarity score distribution, when compared to movie subtitles as input. Figure~\ref{fig:subtitle_vs_screenplay_sentiment} presents the results in terms of different input types. When compared to negative sentiment scores, we observe that the distribution of emotions in subtitles is generally wider, with the median and quartile values being higher than those of the screenplays. This is partly attributed to the dialogues in subtitles being more emotionally charged and lacking contextual connections. In comparison of the neutral sentiment scores, we find that the screenplay shows a significantly higher median than subtitles, indicating that it serves as a more concentrated distribution with less fluctuation. In contrast, we find that using subtitles as input leads to more fluctuating sentiment scores, suggesting a higher possibility of extreme sentiment polarisation. 
Figure~\ref{fig:subtitle_vs_screenplay_sentiment}  shows that positive sentiment fluctuation in subtitles is greater, and screenplays tend to be more stable and concentrated. It indicates that subtitles are more prone to generating emotionally expressive and strongly positive reviews, and screenplays tend to yield more controlled and balanced emotional output.

\subsection{Emotion Analysis}


We use a DistilRoBERTa-based emotion detection model ~\cite{Hartmann2022} to identify the emotions expressed in movie reviews. We categorise the emotions into seven common emotion types: joy, sadness, anger, fear, surprise, disgust, and neutral. We use the model to determine the top five emotions present, along with their corresponding probability distributions.

 As shown in Figure~\ref{fig:emotion-scores-by-question}, we present the LLM-generated movie reviews and calculate the emotion scores for each review under each question (Question 1, Question 2, and Question 3). Similarly, to further examine how different types of inputs affect the emotion analysis results, we also add the movie screenplay as inputs. Following this processing, we obtained the corresponding emotion analysis results, which served as the basis for subsequent visualisation and comparative analysis.

\noindent
\begin{figure*}[!t]
    \centering
    \includegraphics[width=1\linewidth]{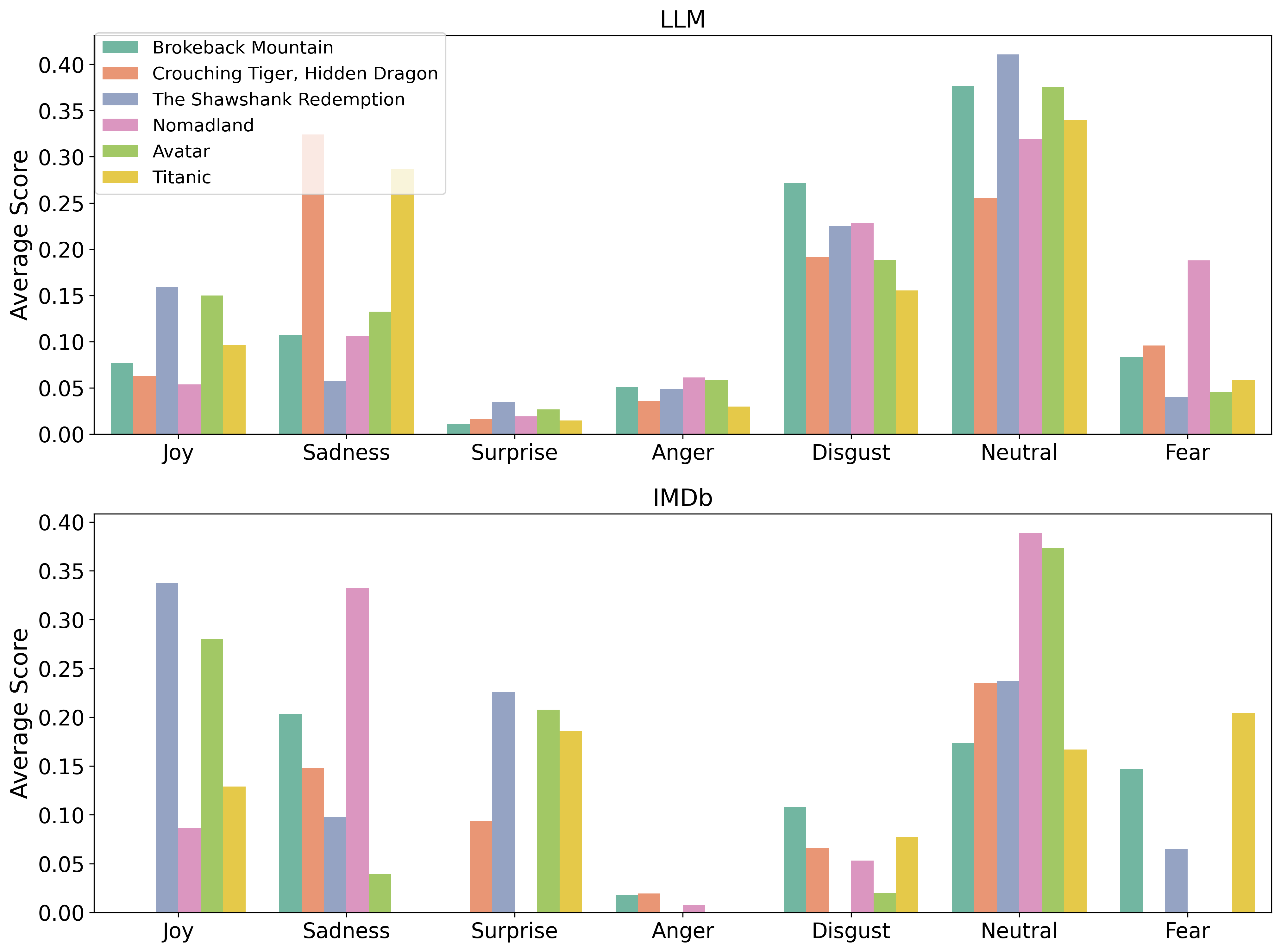}
    \caption{Comparison of emotion scores across different movies between LLM-generated reviews and IMDb reviews.}
    \label{fig:emotion-scores-ai-vs-imdb-movie}
\end{figure*}

\begin{figure*}[!t]
    \centering
    \includegraphics[width=0.95\linewidth]{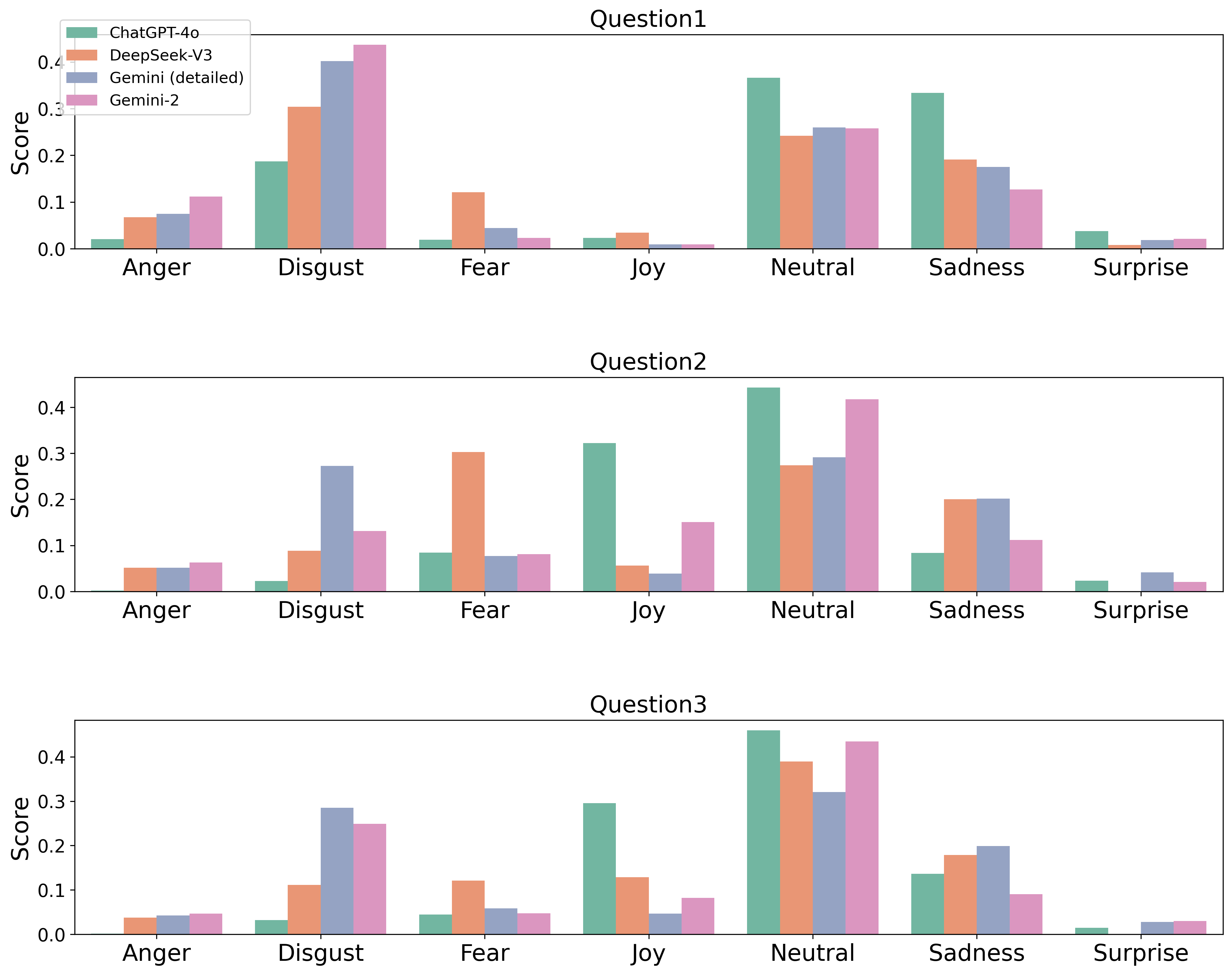}
    \caption{Emotion score by LLMs for Q1 (Negative), Q2 (Positive), and Q3 (Neutral) categories.}
    \label{fig:emotion-scores-by-question}
\end{figure*}

\begin{figure*}[!t]
    \centering
    \includegraphics[width=0.95\linewidth]
    {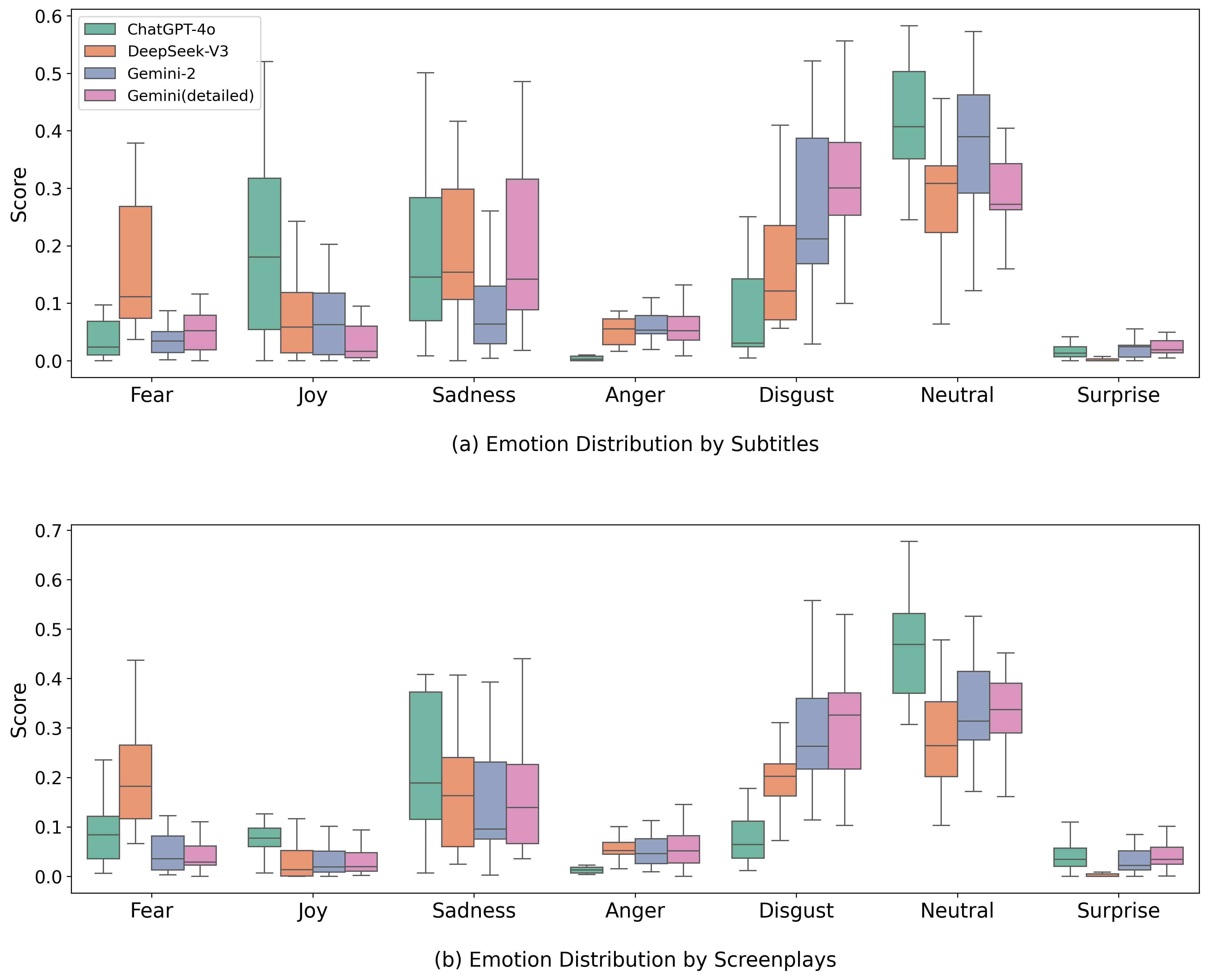}
    \caption{Emotion scores from subtitles and screenplays across LLMs.}
    \label{fig:emotion-sub-vs-script}
\end{figure*}

\begin{table*}[!t]
\centering
\small
\renewcommand{\arraystretch}{1.2}
\resizebox{\textwidth}{!}{
\begin{tabular}{lccccccc}
\hline
\textbf{Model} & \textbf{Surprise} & \textbf{Anger} & \textbf{Neutral} & \textbf{Disgust} & \textbf{Sadness} & \textbf{Fear} & \textbf{Joy} \\
\hline
GPT-4o & 0.033 & 0.015 & 0.156 & 0.167 & 0.228 & 0.052 & 0.376 \\
DeepSeek-V3 & 0.040 & 0.039 & 0.328 & 0.214 & 0.177 & 0.110 & 0.091 \\
Gemini-2.0 & 0.026 & 0.060 & 0.330 & 0.306 & 0.137 & 0.028 & 0.104 \\
Gemini (detailed) & 0.043 & 0.072 & 0.250 & 0.346 & 0.194 & 0.059 & 0.024 \\
IMDb & 0.178 & 0.015 & 0.263 & 0.065 & 0.164 & 0.139 & 0.208 \\
\hline
\end{tabular}}
\caption{Emotion score distribution by various LLMs.}
\label{tab:emotion_subtitles}
\end{table*}


 Table~\ref{tab:emotion_subtitles} compares the performance of various LLMs, where GPT-4o exhibits the most positive emotion, with a high Joy score (0.376), followed by sadness and neutral. This indicates that the overall emotion of the generated reviews leans positive. In contrast, DeepSeek-V3 shows the largest proportion of Neutral emotion (0.328), followed by Disgust and Sadness, respectively, reflecting predominantly neutral and  negative emotions. Meanwhile, Gemini-2.0 and Gemini (detailed) lean towards negative emotions, with higher proportions of disgust and neutral but lower joy.
 
 These findings align with our earlier results of Sentiment Polarity Analysis (Figure 2), reinforcing that  different LLMs have varying preferences in their emotional tendencies and expressive styles \cite{sun2023}. Table~\ref{tab:emotion_subtitles}  shows that GPT-4o is better suited for tasks requiring strong, clearly positive emotional expression, while DeepSeek-V3 is more appropriate for objective, neutral assessments.  Furthermore, Gemini-2.0 and Gemini(detailed) are weaker and more prone to deviations from the expected emotional outcome. Additionally, according to Table \ref{tab:emotion_subtitles}, compared to IMDb user reviews, LLM-generated reviews exhibit more extreme emotional patterns, particularly in the expression of negative emotion (such as sadness and disgust). In contrast, IMDb reviews demonstrate a more balanced distribution across different emotion categories.

We next analyse the differences in emotional tendencies of LLM-generated reviews and then compare the emotion scores between LLM-generated reviews and IMDb user reviews across different movies.  Figure~\ref{fig:emotion-scores-ai-vs-imdb-movie} presents the results for movies including Crouching Tiger and Titanic that exhibit strong negative sentiments, with high sadness scor es and the sum of negative and neutral emotions (exceeding 0.6). In contrast, movies such as The Shawshank Redemption and Avatar show more positive and neutral emotions, with relatively high joy and neutral scores. In these cases, IMDb user reviews express clear positive emotions. Furthermore, the reviews for Titanic and Brokeback Mountain reveal more complex and ambiguous emotional profiles, leaning overall towards neutral and negative sentiments. These examples demonstrate that emotion analysis can intuitively quantify the structure of complex and ambiguous emotions~\cite{Chu2017, Thao2019}.

Figure~\ref{fig:emotion-scores-ai-vs-imdb-movie}  compares the emotion scores among movies between LLM-generated reviews and IMDb user  reviews. We observe that the LLM-generated reviews generally show more extreme disgust and neutral scores than IMDb reviews. In the case of movies such as Brokeback Mountain, Crouching Tiger and The Shawshank Redemption, the disgust scores are noticeably higher than those in IMDb reviews. Furthermore, the neutral scores for Brokeback Mountain and The Shawshank Redemption in LLM-generated reviews are also higher than those in IMDb reviews. These patterns again validate that LLMs tend to amplify emotional intensity when generating reviews.

We compare the performance of LLM guidance under different demands (Figure~\ref{fig:emotion-scores-by-question}), namely negative reviews (Q1), positive reviews (Q2) and neutral reviews (Q3). In Question 1, we find that the disgust sentiment is dominant among various LLMs, with neutral and sadness also being relatively prominent. In Questions 2, only  GPT-4o's joy score is relatively high, while other LLMs concentrate more scores on neutral categories. In Questions 2, all LLMs performs well. In general, we observe that all LLMs respond well to negative prompts. However, only GPT-4o effectively captures positive emotions under explicit prompts. Under neutral prompts, all LLMs generally can capture the tendency.

Furthermore, based on the box plot in Figure~\ref{fig:emotion-sub-vs-script}, we examine how different types of review inputs (subtitles vs screenplays) to the respective LLMs influence the expression of seven distinct emotions in the emotion analysis. We select the most representative emotions (joy, sadness, disgust) for analysis.  We observe that in the case of 'joy', the median values among LLMs are similar, and  GPT-4o, in particular, exhibits a higher median and wider interquartile range based on the subtitle as LLM input. This indicates that subtitles are more likely to generate extreme joy emotions. We observe the same trend for the sadness emotion, where the range of the subtitles' emotion score is notably wider than the rest for DeepSeek-V3 and GPT-4o. This suggests that inputs based on subtitles are more likely to generate reviews with high fluctuations in emotions. Additionally, in screenplay input,  GPT-4o shows wider interquartile ranges for neutral and surprise emotions, indicating more dispersed outputs. In summary, we observe that generating movie reviews based on subtitles results in more extreme and fluctuating expressions of emotions compared to screenplays, which could be due to additional context information of events offered in the screenplays, rather than only dialogues given in subtitles.

\subsection{Similarity Analysis}

 Based on the sentiment analysis described above, we conduct text (review) similarity analysis to evaluate the similarity between generated reviews and human-written reviews from a semantic perspective. We employed cosine similarity at the lexical level as an approximate measure of semantic similarity. Although this method does not capture deep semantic meaning, it effectively reflects the closeness in word usage and structure by comparing the angles between text vector representations.

\begin{figure*}[!t]
    \centering
    \includegraphics[width=0.95\linewidth]{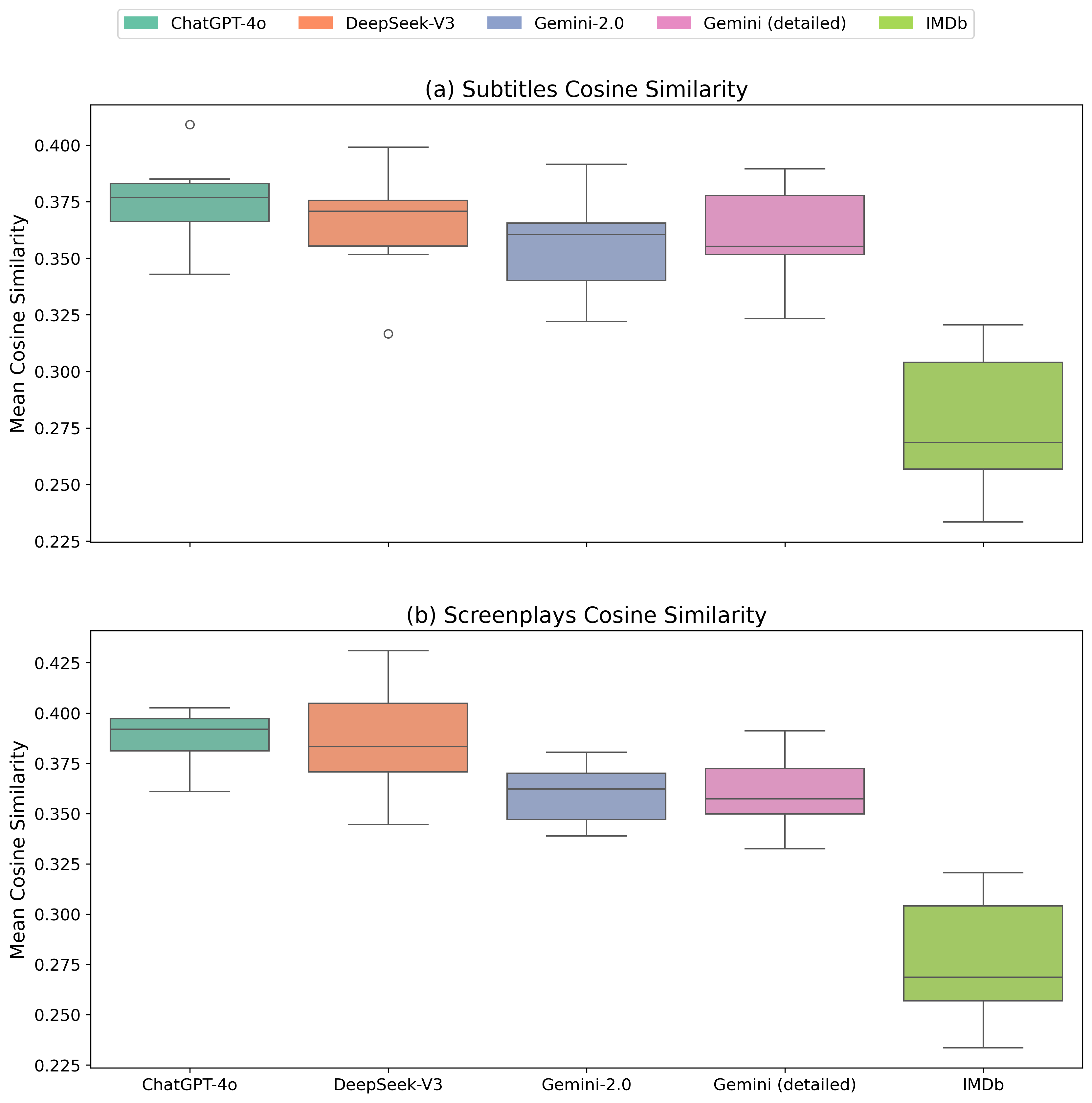}
    \caption{Comparison of Mean Cosine Similarity Scores Across Models Using Subtitles and Screenplays}
    \label{fig:boxplot}
\end{figure*}

\begin{figure*}[!t]
    \centering
    \includegraphics[width=0.95\linewidth]{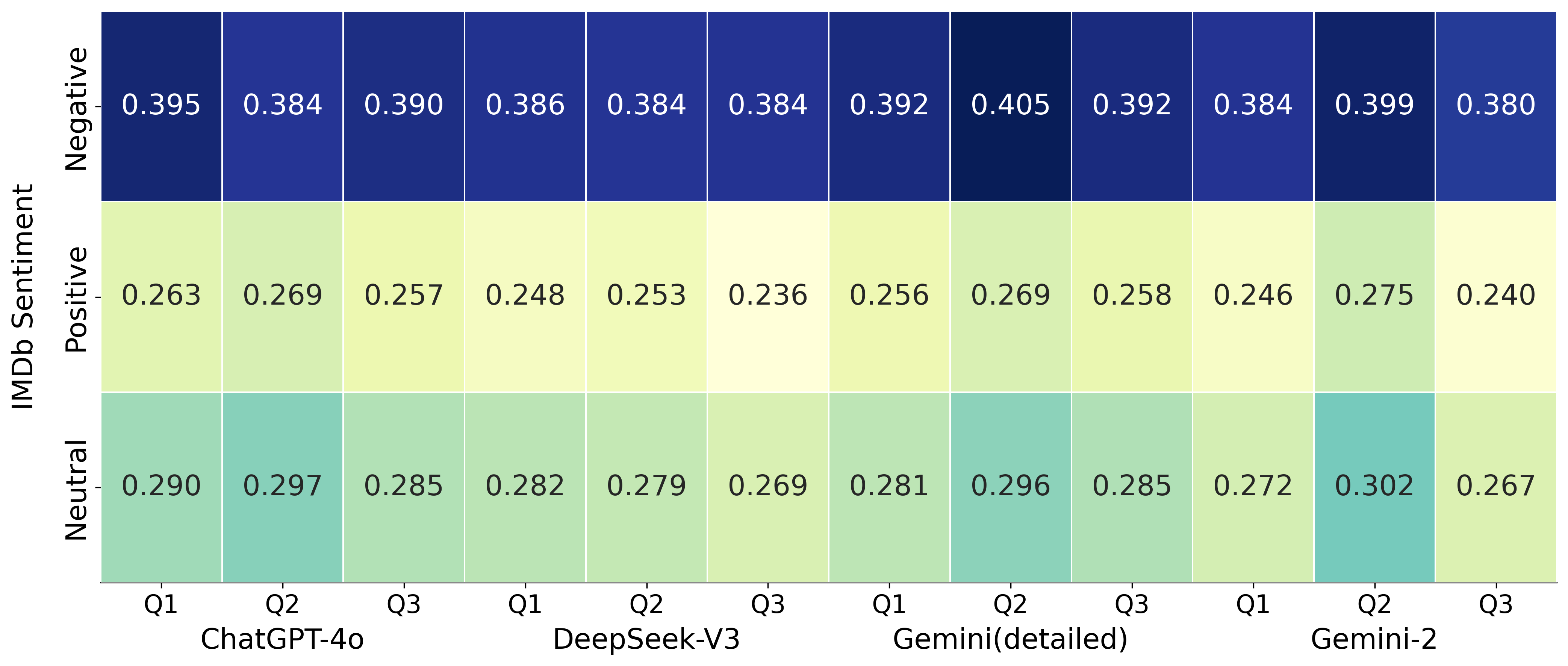}
    \caption{Cosine Similarity by sentiment and Question type for movie subtitle-based reviews for respective LLMs}
    \label{fig:heatplot_subtitles}
\end{figure*}

\begin{figure*}[!t]
    \centering
    \includegraphics[width=0.95\linewidth]{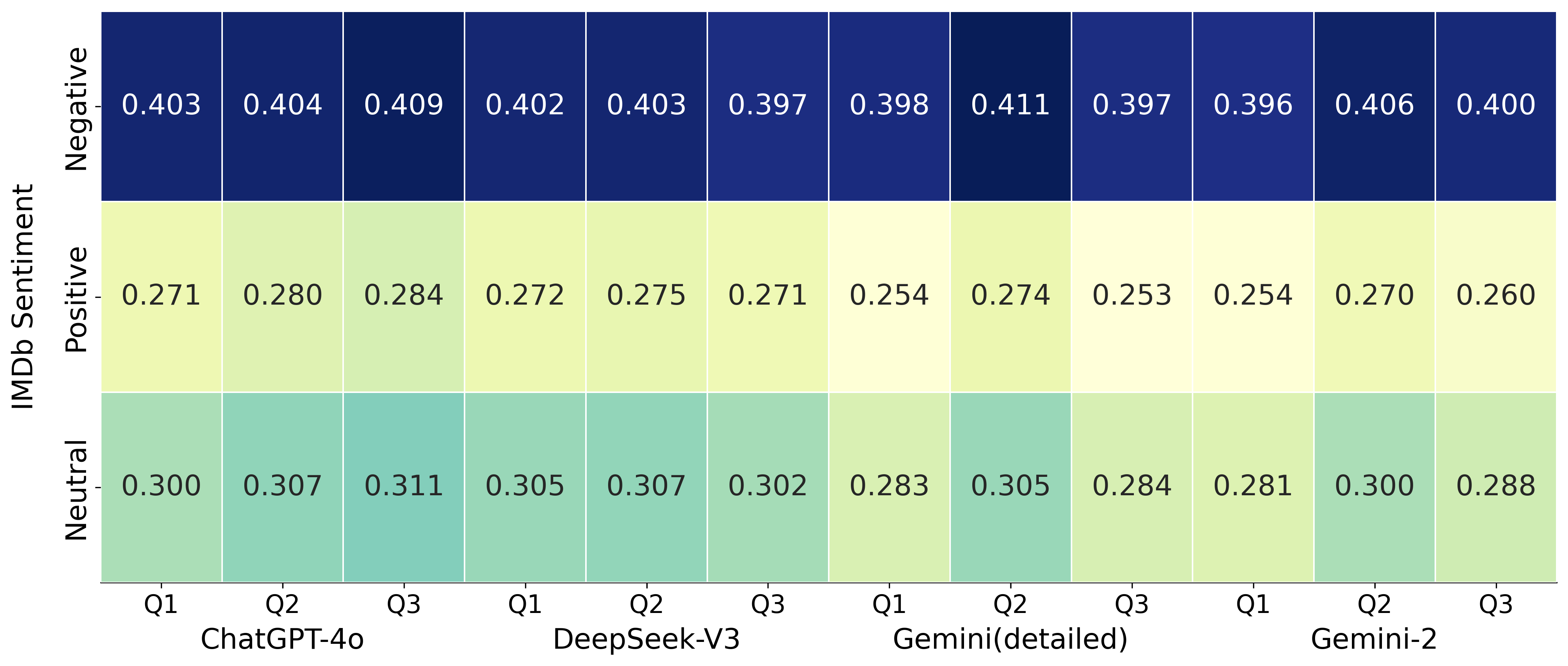}
    \caption{Cosine Similarity by sentiment and Question type for movie screenplay-based reviews for respective LLMs.}
    \label{fig:heatplot_screenplays}
\end{figure*}

 We compute pairwise cosine similarity scores between LLM-generated reviews and IMDb user reviews to evaluate their lexical similarity. We transform each review into a TF-IDF vector to capture word importance, and compute cosine similarity between these vector representations. Although this method does not capture detailed semantics, it effectively measures surface-level textual similarity, allowing us to assess how closely the generated reviews resemble human-written ones in terms of word usage and structure.  Figure~\ref{fig:boxplot} presents the results of this evaluation, while Figures~\ref{fig:heatplot_subtitles} and~\ref{fig:heatplot_screenplays} compare reviews generated from screenplays and subtitles across different sentiment polarities (positive, neutral, negative).

 We examine the internal similarity among IMDb reviews for each movie to analyse the consistency and diversity within IMDb user reviews as given in Figure~\ref{fig:boxplot}. The generally low similarity scores indicate a significant variation in how viewers express their opinions, highlighting the richness of human perspectives. 
 
 Next, for each LLM model, we compare every generated review with its corresponding IMDb user reviews for the same movie, as shown in Figure~\ref{fig:boxplot}. Across all models and generation methods (screenplays or subtitles), the average similarity scores consistently rank in the following order: 1. GPT-4o, 2.  DeepSeek-V3, 3. Gemini instances (Gemini 2.0 and detailed). GPT-4o achieves the highest median similarity and the smallest Inter Quartile Range (IQR), indicating that its generated content is not only of higher quality but also more consistent and stable. DeepSeek-V3 also reaches a relatively high median similarity, but with a noticeably larger IQR than GPT-4o, suggesting greater variability in its output—performing well on some samples while falling short on others. The Gemini instances rank lowest in both median similarity and overall distribution, revealing a generally weaker semantic alignment with IMDb-style reviews under both input conditions.

 As shown in Figures~\ref{fig:heatplot_subtitles} and \ref{fig:heatplot_screenplays}, we can observe that the negative reviews generated by the three LLMs show a higher similarity to IMDb reviews than neutral and positive sentiments. This may result from the tendency of negative reviews to rely on concrete details and specific criticisms, while positive reviews are often more subjective and stylistically diverse.

In Figure~\ref{fig:boxplot}, we can also observe that reviews generated from screenplays are generally more similar to IMDb reviews than those generated from subtitles. This suggests that screenplay-based LLM reviews are more aligned with IMDb user reviews in terms of structure and content. One possible reason is that screenplays typically provide richer contextual information, including character dialogues, narrative progression, and emotional cues, which help the model develop a more comprehensive understanding of the film and generate reviews with greater depth and expressiveness. In contrast, subtitles are more fragmented and lack scene-level and character background information, which may constrain the model’s ability to capture and evaluate the full storyline.

We further analyse the effect of different contextual LLM prompts on review generation using the Gemini-2.0 model, comparing standard versus detailed persona prompts across screenplay and subtitle inputs. Five detailed persona prompts were introduced: (1) a sensitive and insightful young female movie fan, (2) a sharp and experienced professional critic, (3) a fast-talking, humorous high school movie club president, (4) an angry, critical right-wing male reviewer, and (5) a witty but controversial mature female audience member. The persona prompts also play a significant role in influencing the semantic quality of generated reviews, as shown in Figure ~\ref{fig:heatplot_subtitles}. Comparing the performance of Gemini-2.0 and Gemini (detailed) across both input conditions, we observe that although the standard persona prompts exhibit slightly higher median similarity, the detailed persona prompts demonstrate clear advantages in output stability and robustness against extreme cases. Specifically, the reviews generated with detailed persona prompts display a more concentrated distribution and a smaller interquartile range, indicating more consistent and less variable outputs. This suggests that detailed persona prompts can effectively suppress low-quality review generations and thereby improve the overall reliability of model outputs. In contrast, while the standard persona prompts achieve slightly higher typical performance, they are associated with greater variability and an increased risk of generating lower-quality content.

\subsection{Quantitative Analysis}

Finally, we designed a public survey for qualitative analysis of LLM reviews. We randomly select five reviews from the complete set of IMDb reviews and those generated by three different LLMs. We ask the following four questions using an online survey: 

\begin{enumerate} \item Do you think this review was written by a human or generated by LLMs? \item Do you think the sentiment of this review is positive, negative, or neutral? \item What emotion do you think best describes the tone of this review? \end{enumerate}

We received a total of 50 responses, and the statistical results are shown in the table below.

\begin{table*}[t]
\centering
\renewcommand{\arraystretch}{1.2}

\begin{tabular}{|c|c|c|}
\hline
\textbf{Review Number} & \textbf{Source} & \textbf{Accuracy (\%)} \\
\hline
Review 1 & IMDb &  44 \\
\hline
Review 2 & Gemini & 72 \\
\hline
Review 3 & Gemini &  22 \\
\hline
Review 4 &  GPT-4o &  36 \\
\hline
Review 5 & IMDb &  80 \\
\hline
\end{tabular}
\caption{Accuracy of identifying whether each review was written by a human or generated by LLM.}
\label{tab:review_accuracy}
\end{table*}

\begin{table*}[t]
\centering
\renewcommand{\arraystretch}{1.2}
\begin{tabular}{|c|c|c|c|c|}
\hline
\textbf{Review} & \textbf{Positive} & \textbf{Neutral} & \textbf{Negative} & \textbf{Dominant Sentiment} \\
\hline
Review 1 & 31 & 14 & 5  & Positive \\
\hline
Review 2 & 21 & 26 & 3  & Neutral \\
\hline
Review 3 & 5  & 6  & 39 & Negative \\
\hline
Review 4 & 32 & 16 & 2  & Positive \\
\hline
Review 5 & 44 & 5 & 1  & Positive \\
\hline
\end{tabular}
\caption{Sentiment polarity distribution for each review.}
\label{tab:sentiment_distribution}
\end{table*}

\begin{table*}[t]
\centering
\renewcommand{\arraystretch}{1.2}

\begin{tabular}{|c|c|c|}
\hline
\textbf{Review} & \textbf{Dominant Emotion} & \textbf{Count} \\
\hline
Review 1 & Sadness & 16 \\
\hline
Review 2 & Joy     & 22 \\
\hline
Review 3 & Sadness & 14 \\
\hline
Review 4 & Disgust & 10 \\
\hline
Review 5 & Joy     & 22 \\
\hline
\end{tabular}
\caption{Dominant emotion and its frequency for each review.}
\label{tab:dominant_emotion_summary}
\end{table*}

Table~\ref{tab:review_accuracy}, among the two reviews sourced from IMDb, Review 5 has an accuracy rate of 0.8. This review is structurally clear and has natural emotions, making it relatively easy to be identified as an IMDb movie review~\cite{zhang2023}. In contrast, Review 1 has an accuracy rate of only  44 percent, probably because it uses first-person narration and plain language, which is similar to LLM-generated reviews, making it more difficult to distinguish. Among the three LLM-generated reviews, Review 2 has the highest accuracy rate at  72 percent. This may be attributed to phrases such as “Okay, here’s my review...”, which resemble an LLM-generated tone. However, Review 3 has been only correctly identified as LLM-generated by 22 percent, this could be due to its more colloquial and emotionally expressive tone, which makes it easier to mistake for an IMDb review.

Table~\ref{tab:sentiment_distribution} illustrates that the strong negative sentiment of Review 3, with a negative proportion of 39 out of 50, again validates that Gemini-2.0 tends to generate more extreme negative content under such prompts. This reinforces the conclusion that LLM-generated reviews generally amplify emotions like sadness and disgust. Review 2, generated by GPT-4o, displays a relatively strong expression of joy. This again validates the previous conclusion in the Sentiment Analysis section: GPT-4o performs well in expressing positive emotions. On the other hand, the positive sentiment recognition ratio of Review 5 is 44 out of 50. As shown in Table~\ref{tab:review_accuracy}, the correct identification rate of Review 5 as an IMDb review is also the highest. This suggests that positive sentiment is more consistently recognised across both human-written and LLM-generated reviews, with positively worded reviews tending to exhibit clearer emotion, resulting in higher accuracy in sentiment recognition.

As shown in Table~\ref{tab:dominant_emotion_summary}, Review 2 and 5 display a relatively strong expression of joy. However, Review 2 shows a different dominant emotion compared to its sentiment polarity classification in Table~\ref{tab:sentiment_distribution}. The expression of positive emotion is not as strong, reflecting a stronger tendency of emotion controlled by the prompt. The two Gemini-2.0-generated reviews exhibit different emotions, which reveal the flexibility or instability of LLM's emotional expression~\cite{birhane2023}. The questionnaire results align with the emotion scores, as shown in the previous Table~\ref{tab:emotion_subtitles}, where disgust and sadness have relatively higher scores. This suggests that Gemini-2.0 tends to generate content with more negative emotional tones, and this style is easily distinguished by users.

The questionnaire results further validate the conclusions from the Sentiment Analysis: GPT-4o tends to express positive emotions more effectively, while Gemini-2.0 performs well under negative prompts. Reviews with a clear structure and natural emotional flow are more likely to be recognised as IMDb reviews. In contrast, reviews with exaggerated or inconsistent emotions, especially under negative prompts or subtitle inputs, are more likely to be questioned. This highlights that emotional coherence and moderation are key to generating believable and human-like reviews~\cite{zhang2023, clark2021}.

\section{Discussion}

We generated movie reviews using prominent 
LLMs (GPT-4o, Gemini-2.0, and DeepSeek-V3) and used four methods, including n-gram, sentiment polarity analysis, emotion analysis, cosine similarity analysis, and a questionnaire to systematically compare their performance.  Our focus was on the analysis of LLM reviews and how they differ in emotional expression, language style, and whether humans can identify LLMs from human-written reviews.  In summary, we observed that LLMs are capable of generating reviews that are comparable to IMDB user reviews.

 The results of the emotional analysis (Figure~\ref{fig:polarity-ai-imdb-comparison}, Figure~\ref{fig:emotion-scores-ai-vs-imdb-movie}, and Figure~\ref{fig:emotion-scores-by-question}) show that GPT-4o tends to express more positive emotions and also aligns more closely with IMDb review. In contrast, Gemini instances are more likely to generate emotions such as "disgust" and "sadness" under negative prompts. We found that DeepSeek-V3 offers more balanced and neutral emotions. Our sentiment polarity analysis further supports this trend, with  GPT-4o showing significantly higher positive polarity scores than the other models. The trigram analysis shows that LLM-generated reviews often lack character names and use more \textit{templated language}, revealing stylistic differences from IMDb user reviews.  The Semantic Similarity Analysis (Figure~\ref{fig:boxplot}, Figure~\ref{fig:heatplot_subtitles}, and Figure~\ref{fig:heatplot_screenplays}) shows that reviews generated by GPT-4o are semantically the closest to IMDb user reviews, while those generated by Gemini instances and DeepSeek-V3 exhibit more distinct and individual expression styles.  The results of the questionnaire (Table~\ref{tab:review_accuracy}, Table~\ref{tab:sentiment_distribution} and Table~\ref{tab:dominant_emotion_summary}) further confirmed the findings from sentiment polarity and emotion analysis. The reviews with clear structure and natural emotional flow are more likely to be seen as written by humans, while templated or emotionally inconsistent reviews are more often judged as LLM-generated. These results reveal stylistic differences and content preferences among different language models in generating movie reviews.

In the comparison between Gemini-2.0 and Gemini (detailed), we find that prompting strategies have a significant impact on the quality of generated reviews. While standard prompts achieve slightly higher median performance, detailed prompts offer greater stability and consistency, effectively reducing low-quality outputs and improving the overall reliability of generation. Some of our findings echo previous studies and further reveal how different input types (such as subtitles and screenplays) and prompt styles influence emotional expression, as well as the perceived naturalness of model-generated content. We observed that  GPT-4o exhibits more pronounced positive emotion, with significantly higher scores in the “joy” dimension compared to other models. This trend aligns with the analysis of Hadi et al. \cite{hadi2023}, who noted that mainstream models like  GPT-4o are often optimised to produce neutral or positive outputs to ensure safety and user acceptance, thereby avoiding negative or offensive language. In contrast, Gemini-2.0 was more prone to generating strong expressions of “disgust” and “sadness” in response to negative prompts. Although  the study by Chandra and Ren \cite{chandra2025} focused on movie dialogues, they also pointed out that abusive contexts tend to involve more extreme and emotionally charged language, a pattern that resonates with our observations in LLM-generated reviews under similar conditions.

Moreover, this study employs full-length subtitles and screenplays as input, extending prior generation tasks that typically rely on short texts \cite{wu2024longgenbench}. Our design evaluates models’ ability to maintain structure and coherence in long-form content, which aligns with the goals of LongGenBench for assessing structured generation over extended sequences. In terms of input format effects, we further found that subtitle-based reviews tend to be more emotionally volatile and stylistically expressive, while screenplay-based inputs yield more stable and coherent outputs. Similar issues of input sensitivity have also been addressed by Huang et al., who reported that Transformer-based models are highly sensitive to variations in input length and structure, which may result in noticeable shifts in output style and consistency \cite{huang2023}. These findings provide new empirical support for understanding how LLMs perform in terms of emotional authenticity and linguistic naturalness, while also deepening our insights into how input conditions shape generation style.

Despite adopting a systematic approach in data sourcing, analytical framework design, and model selection, this study still encountered several limitations. First, in terms of sample selection, we analysed only six representative Oscar-winning or nominated movies. These movies featured well-developed plot structures and rich emotional expressions, and their review data remained relatively abundant. As a result, the performance of LLMs in generating reviews appeared overstated. Had we used commercial movies, lesser-known titles, or non-English movies instead, the models' performance might have significantly declined-especially under varying cultural backgrounds and linguistic conventions.
Second, although we applied persona prompting to assign specific character settings and expressive styles to the models, all three models (GPT-4o, Gemini-2.0, and DeepSeek-V3) enforced content moderation mechanisms. These mechanisms often produced less authentic negative expressions~\cite{zhang2023sentiment}, which affected the realism and depth of the generated reviews~\cite{das2023multimodal}. GPT-4o consistently generated predominantly positive content, often softening negative critiques. Gemini-2.0 produced outputs that leaned toward neutrality and relied on evasive language when expressing criticism. DeepSeek-V3 displayed a greater willingness to deliver critical language, but frequently lacked emotional variety, which resulted in structured and templated review patterns. Third, although we employed emotion analysis tools such as DistilRoBERTa (fine-tuned for this task), these tools still misinterpreted highly subjective and context-rich movie reviews. Movies like \textit{Nomadland} and \textit{Brokeback Mountain}, which conveyed complex emotions and layered metaphors, often inspired reviews that intertwined sentiments such as hope amidst sorrow or understanding within anger. These nuanced expressions posed significant challenges for emotion classification models.

Moreover, as this study only compared standardised and detailed persona prompts using the Gemini-2.0 model, the generalizability of these findings to other LLMs remains uncertain. Different AI models may exhibit varying sensitivities to input structure and may generate different patterns of responses under the same conditions. Although our observations provide a plausible inference, further experiments involving diverse models are needed to validate whether this trend applies universally across LLM architectures. Finally, during the qualitative evaluation involving human participants, we designed a questionnaire to assess the emotional accuracy of LLM-generated reviews and their ability to "fool" readers. However, the sample size remained limited. Additionally, the participants came from diverse backgrounds with varying standards of evaluation and reading comprehension, which likely affected the representativeness of the statistical conclusions.

Although this study demonstrated the potential of LLMs in generating movie reviews, we also identified several challenges, including unnatural emotional expression, incomplete information retrieval, and the limited generation of negative comments. Future research could address these issues in the following ways. Current mainstream LLMs generated overly conservative or evasive expressions when handling negative sentiments. Future work could incorporate high-quality, authentic negative review corpora during training, especially datasets rich in strong emotions, detailed descriptions, and critical language. This approach could improve the models’ ability to reason about and express negative emotions such as anger, disgust, and sadness.

This study relied solely on textual inputs such as subtitles and screenplays, while real-world movie reviews often required multimodal understanding~\cite{tsimpoukelli2021}. Future research could integrate video clips, audio, and cinematographic language into the generation pipeline~\cite{yang2024large}. Multimodal deep learning \textcolor{blue}{~\cite{jabeen2023review}} that combines visual, textual, and auditory data-could help LLMs better interpret character actions, cinematographic styles, and emotional atmospheres. For example, intense fight scenes, anxiety induced by rapid camera cuts, emotionally charged musical cues, or symbolic colour schemes all represented elements that human reviewers frequently emphasised. As a cultural medium, movies reflected specific cultural~\cite{myung2024blend}, historical, regional, gender, and class contexts, which also shaped the language of reviews~\cite{pawar2024survey}. In our study, we gathered that LLMs typically lacked  background of cultural knowledge and future studies could integrate external databases rich in contextual and cultural information, enabling models to incorporate these references automatically. Therefore, LLMs for cinema and television can be developed with specialised training data. Furthermore, building multilingual movie review corpora could improve LLMs' ability to generate reviews in languages such as Chinese, French, and Spanish, thus improving their applicability in the global movie market~\cite{etxaniz2024bertaqa}.

Existing automatic evaluation methods, such as sentiment scoring and cosine similarity, offered objective metrics but struggled to assess how emotionally resonant or thought-provoking a review appeared. Future research could introduce more complex evaluation frameworks by combining expert ratings or crowd-sourced assessments. This approach could help LLMs continuously improve the quality of generated reviews and align them more closely with both user experiences and professional review standards~\cite{hada2023large}. There are major challenges in analyzing movie narrative structures, thematic evolution, and long-term emotional feedback from audiences, especially when combining multimodel data (such as text, audio, and video) for comprehensive analysis~\cite{raiaan2024}.

The main contributions of this study include:
\begin{itemize}
    \item Genre adaptability: Exploring whether LLMs can generate consistent reviews for different types of movies.
    \item Emotional consistency and bias: Analyzing whether LLM-generated reviews can maintain consistent emotional expression and identify potential bias issues.
    \item Text quality and readability: Evaluating the performance of LLM-generated reviews in terms of fluency, emotional depth, and informational value.
\end{itemize}


\section{Conclusion}

This study systematically compared movie reviews generated by three LLMs, including GPT-4o, DeepSeek-V3, and Gemini-2.0 and compared them with IMDb user reviews, focusing on linguistic features, emotional alignment, and semantic similarity. 

Sentiment polarity and emotion analysis confirmed that LLMs can effectively distinguish between positive, neutral, and negative review prompts when clearly instructed. The emotional tendencies varied across LLMs:  GPT-4o leaned toward positive sentiment, generally performs well under specific prompts. DeepSeek-V3 maintained balanced and neutral emotions and aligned well with IMDb reviews, while Gemini-2.0 expressed more ambiguous and mildly negative emotions. Compared to IMDb user reviews, LLM-generated reviews sometimes amplify emotions. Additionally, the choice of input source significantly influenced emotional variance — reviews based on subtitles exhibited more polarised and extreme sentiment, while those based on screenplays resulted in more stable and consistent emotional output. The text similarity analysis showed that reviews generated by GPT-4o were semantically closest to IMDb reviews, while those produced by Gemini instances, though more diverse in language style, were structurally less aligned with IMDb content. The reviews with negative sentiment prompts tended to achieve higher similarity with IMDb reviews compared to those with positive or neutral sentiment. Additionally, screenplay-based reviews exhibited higher cosine similarity scores than subtitle-based ones. Detailed persona prompts led to more expressive and nuanced outputs.

Our study highlighted both the strengths and limitations of LLMs in generating movie reviews based on movie subtitles and screenplays. Although LLMs demonstrated strong potential for producing grammatically sound, sentiment-aligned, and coherent reviews, they still face challenges in mimicking the emotional depth, subjective nuance, and detail-oriented style found in IMDb reviews. Future research should focus on enhancing emotional control, expanding model training with diverse datasets, and incorporating multimodal inputs to help LLMs better capture the richness of IMDb review writing.


\section*{Code and Data Availability}
Our code and data are available via our GitHub repository:
\url{https://github.com/sydney-machine-learning/LLM-moviereviews.git}

\section*{Declarations}
The authors declare that they have no known competing financial interests or personal relationships that could have appeared to influence this study.

\section*{Author Contributions }

Brendan Sands contributed to coding, design and implementation, analysis of results, visualisation and editing. Chenhao Xu contributed to writing the original draft, visualisation, and quantitative analysis.
Yining Wang contributed to data analysis, literature review, questionnaire design, writing (original draft), and subsequent editing and revision. Yuxuan Zhou contributed to the literature review, writing and revisions, visualisations and data analysis. 
Lai Wei contributed to data collection, coding, data cleaning, visualisation and analysis  of results.

R. Chandra contributed to the conceptualisation, project supervision, editing of the final draft, experiment design, and analysis. 

\bibliographystyle{IEEEtran}

\bibliography{refs}

\appendix

\appendix
\section*{Appendix A: Detailed Prompt Templates for LLM-Generated Movie Reviews}
\label{appendix:prompts}

The following detailed prompt templates are the alternative, more detailed prompt inputs used to guide Gemini 2.0 in generating the alternative Gemini-detailed movie reviews. 

\begin{enumerate}
  \item \textbf{Young Professional Woman (Late 20s)}

  I have subtitle text from a movie. Please read the subtitles and generate a short movie review written in the voice of a young professional woman in her late 20s who loves going to the movies and has a sharp, thoughtful perspective. She often posts casual but insightful reviews on Letterboxd. Her tone is witty, self-aware, and a little emotionally vulnerable. Write a review of this movie from her point of view, focusing on the emotional beats, themes, characters, the actor performances, and overall experience of watching the film. Make it sound like something she might post online the night after seeing it.

  \item \textbf{Professional Film Critic (Late 60s)}

  I have subtitle text from a movie. Please read the subtitles and generate a short movie review written in the voice of a professional film critic in their late 60s who has a dry, to the point, and somewhat cynical perspective. Their tone is witty, self-aware, and cuts to the point. Write a review of this movie from their point of view, as if they just watched the movie, focusing on the emotional beats, themes, characters, the actor performances, and overall experience of watching the film. Make it sound like something they might post in an IMDb review after seeing it.

  \item \textbf{High School Film Club President (Age 17)}

  I have subtitle text from a movie. Please read the subtitles and generate a movie review written in the voice of a 17-year-old high school student who runs their school’s film club. They’re very online, love A24 movies, and aren’t afraid to speak their mind. Their tone is fast-paced, funny, and packed with pop culture references, while still showing an impressive grasp of film analysis. Write a review of this movie from their point of view, as if they just watched the movie, mixing humor with surprisingly sharp takes on themes, symbolism, and character arcs. Feel free to comment on the actor performances.

  \item \textbf{Right-Wing Young Man (Age 30s)}

  I have subtitle text from a movie. Please read the subtitles and generate a short movie review written in the voice of a young right-wing man in his 30s. His tone is slightly angry, and he is a little aggressive. He rarely gives positive reviews, preferring to focus on the negative. Provide a review as if you had just been to see the movie. You express comments on the actors, characters, and the quality of the plot.

  \item \textbf{Older Professional Woman (Late 40s)}

  I have subtitle text from a movie. Please read the subtitles and generate a short movie review written in the voice of an older professional woman in her late 40s who loves going to the movies and has a thoughtful and humorous perspective. Her tone is dry, and a little controversial. Write a review of this movie from her point of view, focusing on the themes, characters, the actor performances, and overall experience of watching the film. Make it sound like something she might post online the night after seeing it.

\end{enumerate}

\section*{Appendix B:  Reviews}

Movie: Nomadland
Movie text: Subtitles
Prompt: Context 1, Question 1. (Bad Review)

\begin{enumerate}
  \item \textbf{GPT4o}

"\textit{Nomadland}" is a film that seems to have captivated audiences with its portrayal of the modern-day nomadic life. However, from my perspective, it failed to deliver in several areas, leading to a rather underwhelming cinematic experience. Firstly, as a narrative,"\textit{Nomadland}" struggles with pacing. The film meanders through various scenes without a clear sense of direction or purpose. It attempts to capture the essence of a transient lifestyle, but in doing so, it loses the engaging momentum typically expected in storytelling. The film feels like a string of vignettes loosely tied together, lacking a cohesive storyline to truly immerse the viewer. Moreover, while Frances McDormand delivers a nuanced performance, the film’s dialogue is often stilted and lacks the depth necessary to emotionally invest in the characters. Conversations between characters feel contrived, as if they're merely vehicles for philosophical musings rather than organic interactions. Visually, the film captures some stunning landscapes, but this aesthetic is not enough to compensate for the lack of substantive content. There's a repetitive quality to the imagery that dilutes its impact over time. Instead of adding layers to the narrative, these scenic shots become fillers in an already prolonged runtime. Finally, while the film aims to shed light on the hidden communities of modern American nomads, it often skirts the deeper systemic issues that force people into this lifestyle. It romanticizes the nomadic experience without fully addressing the hardships and socioeconomic factors that contribute to it. The film misses an opportunity to dive deeper into these aspects, resulting in a superficial exploration of a complex subject. In conclusion,"\textit{Nomadland}" may appeal to some for its contemplative nature and picturesque footage, but those seeking a compelling narrative with rich character development might find it lacking. The film tries to be profound but ends up feeling like a missed opportunity for a more incisive look into the lives it portrays.

\item \textbf{Deepseek-V3}

\textit{Nomadland} Review: A Meandering, Pretentious Slog If you’re looking for a film that mistakes slow, aimless wandering for profound storytelling, \textit{Nomadland} is your movie. Directed by Chloé Zhao, this Oscar-winning drama follows Fern (Frances McDormand), a woman who takes to the road after losing her job and husband, living out of her van while drifting between seasonal work and encounters with other nomads. The Good: Frances McDormand delivers a solid performance, as expected, and the cinematography captures the stark beauty of the American West. Some of the supporting cast-real-life nomads playing fictionalized versions of themselves-bring moments of authenticity. The Bad (And There’s A Lot): - Pacing That Feels Like Eternity: This movie moves at the speed of a dying cactus. Long, lingering shots of Fern staring into the distance or doing mundane tasks don’t add depth—they just test your patience. - Emotionally Hollow: For a film about grief and resilience, \textit{Nomadland} keeps its protagonist (and audience) at arm’s length. Fern’s journey should be moving, but it’s so understated it borders on inert. - Pretentious Vagueness: The script mistakes vagueness for profundity. Lines like "Home-is it just a word, or is it something you carry within you?" sound deep but say nothing. The film wants to be poetic but ends up feeling like a self-indulgent Instagram post about \~wanderlust\~. - No Real Plot: If you enjoy movies where nothing happens-literally nothing-this is for you. Fern drifts, meets people, drifts some more, and… that’s it. No stakes, no growth, just vibes. Final Verdict: \textit{Nomadland} is beautifully shot but emotionally vacant, mistaking stillness for substance. It’s the kind of film critics adore (hence the Oscars) but leaves average viewers wondering, "That’s it?" If you want a meditative character study, maybe it’s for you. If you want a story with momentum, emotion, or purpose, keep driving. Rating: (2/5) – Gorgeous to look at, painful to sit through.

\item \textbf{Gemin 2.0}

Okay, so I just saw \textit{Nomadland}, and honestly? I'm kind of baffled by all the hype. I went in expecting this profound, moving portrait of resilience, and I left feeling like I'd just watched a super long, slightly depressing tourism ad for...living in your van. First of all, the pacing was glacial. Seriously, I think I aged about five years during this movie. Every scene just lingered and lingered. Yes, I get it, life on the road is slow, but that doesn't mean I want to experience it in real-time in a movie theatre. There were so many shots of just…landscapes. Beautiful landscapes, sure, but after a while, they all started to look the same. I felt like I was watching a screensaver for two hours. And Fern, the main character? Frances McDormand is a great actress, I know, but I just couldn't connect with her. She's so stoic and closed off that it's hard to understand her motivations. Why is she so determined to live this nomadic life? Is it truly about freedom, or is it just a way to avoid dealing with her grief and loss? The movie hints at things, but it never really digs deep enough. Plus, the "real people" they cast felt... I don't know, exploitative? Like they were being used as props to add a layer of authenticity. It felt a little condescending, honestly, like the film was saying,"Look at these quirky, down-on-their-luck folks! Aren't they interesting?" And the ending? What ending? It just kind of…stopped. Fern goes back to her abandoned town for a minute and then drives away. The movie doesn't offer any real resolution or insight. Look, I appreciate that \textit{Nomadland} tried to shed light on a different way of life, but it ultimately felt hollow and pretentious. It was beautiful to look at, I will give it that, but it was also incredibly boring, emotionally distant, and ultimately, kind of meaningless. Save your money and go on a road trip yourself - you'll probably have a more engaging experience. Two stars, and that's mostly for the scenery. 

\item \textbf{Gemini 2.0 (Detailed Context)}

Okay, buckle up, buttercups, because I just saw \textit{Nomadland} and, whew, where do I even begin? Let's just say it's not exactly landing on my "best of" list this year. First off, I get the whole "finding yourself" thing, and Frances McDormand is always watchable, even when she's basically playing a sad beige mannequin (sorry, Fern!). But the relentless misery parade? Girl, give us a break! It felt like a two-hour commercial for leaving society behind and embracing the "simple life" in a van. Influencer vibes, but make it depressing. The emotional beats felt so manufactured, like the filmmakers were daring us to cry by throwing every type of loss at us. Husband? Gone. Town? Gone. Job? Gone. Retirement? LOL, good one. It was less "heartfelt journey" and more "poverty tourism" with a side of virtue signaling. And the characters! Oh, honey, they were *characters*. A parade of quirky, weathered folks with "wisdom" to impart, each one seemingly more tragic than the last. I felt like I was trapped in a retirement home that had been swallowed by a dust storm. It's like, I get that life is hard, but did everyone have to be a caricature of rugged individualism? Where's the joy? Where's the nuance? Show me some actual happiness with these people! The whole "community" aspect also felt a little...icky. Like, are we supposed to celebrate these people being forced into this lifestyle by economic circumstances? It felt like romanticizing survival, and I'm not about that life. I can respect their resilience, but to act like it is something to envy is just a bit insulting. Overall, \textit{Nomadland} was a beautiful-looking bummer. I appreciate the attempt at empathy, but it felt more like emotional manipulation disguised as profound insight. I'm all for introspective cinema, but this one left me feeling drained and a little bit exploited. Maybe I'm just basic, but I need a little bit of hope with my heartbreak. Guess I'll go rewatch Paddington 2 to cleanse my soul.

\end{enumerate}

\end{document}